\documentclass[10pt,twocolumn,letterpaper]{article}

\usepackage{cvpr}              

\usepackage{graphicx}
\usepackage{amsmath}
\usepackage{amssymb}
\usepackage{multicol, blindtext}
\usepackage{subcaption}
\usepackage{cite}
\usepackage{soul}
\usepackage{units}
\usepackage{array}
\usepackage{multirow}
\usepackage{caption}
\usepackage{float}
\usepackage{booktabs}
\usepackage{times}
\usepackage{tabularx}
\usepackage{nicefrac}
\usepackage{enumitem}
\usepackage{amssymb}

\usepackage{adjustbox}

\newcolumntype{R}[2]{
    >{\adjustbox{angle=#1,lap=\width-(#2)}\bgroup}%
    l
    <{\egroup}%
}

\newcolumntype{L}[1]{>{\raggedright\arraybackslash}p{#1}}
\newcolumntype{C}[1]{>{\centering\arraybackslash}p{#1}}
\newcolumntype{R}[1]{>{\raggedleft\arraybackslash}p{#1}}
\usepackage{pifont}
\newcommand{\cmark}{\ding{51}}
\newcommand{\xmark}{\ding{55}}

\usepackage{caption}
\usepackage{xcolor}
\usepackage{mathtools}
\usepackage{xfrac}
\usepackage{wasysym}
\usepackage{algorithm}%
\usepackage{algpseudocode}%
\usepackage[accsupp]{axessibility}  

\makeatletter

\@namedef{ver@everyshi.sty}{}
\newcommand\notsotiny{\@setfontsize\notsotiny{6.31415}{7.1828}} 

\makeatother

\usepackage{comment}
\usepackage{pgfplots}
\usepackage{tikz} 
\usetikzlibrary{patterns}
\pgfplotsset{compat=newest}

\usepackage[pagebackref,breaklinks,colorlinks]{hyperref}

\usepackage[capitalize]{cleveref}
\crefname{section}{Sec.}{Secs.}
\Crefname{section}{Section}{Sections}
\Crefname{table}{Table}{Tables}
\crefname{table}{Tab.}{Tabs.}

\newcommand{\probP}{\kern0.15em \text{I\kern-0.15em P}}

\newcommand{\wet}{\textvisiblespace \kern0.25em (ours)}

\newcommand{\both}{\textvisiblespace \kern-0.45em * (ours)}

\definecolor{water_color}{RGB}{69,156,238}
\definecolor{eth_orange}{RGB}{255,126,40}
\definecolor{eth_green}{RGB}{153,204,51}
\definecolor{eth_blue}{RGB}{169,204,242}
\definecolor{eth_red}{RGB}{230,140,132}
\definecolor{eth_gray}{RGB}{126,126,126}

\newcommand{\PAR}[1]{\vspace{-0.2eM}\vskip4pt \noindent{\bf #1}}

\def\cvprPaperID{11286} 
\def\confName{CVPR}
\def\confYear{2024}

\begin{document}

\title{Cross-spectral Gated-RGB Stereo Depth Estimation\vspace{-0.5eM}
}

\author{Samuel Brucker$^1$\qquad Stefanie Walz$^2$\qquad Mario Bijelic$^{1,3}$\qquad  Felix Heide$^{1,3}$\vspace{0.5pt}
\and
\centerline{$^1$Torc Robotics \quad $^2$Mercedes-Benz \quad $^3$Princeton University}}

\maketitle

\begin{abstract}
\vspace{-0.7eM}

Gated cameras flood-illuminate a scene and capture the time-gated impulse response of a scene. By employing nanosecond-scale gates, existing sensors are capable of capturing mega-pixel gated images, delivering dense depth improving on today's LiDAR sensors in spatial resolution and depth precision. Although gated depth estimation methods deliver a million of depth estimates per frame, their resolution is still an order below existing RGB imaging methods. In this work, we combine high-resolution stereo HDR RCCB cameras with gated imaging, allowing us to exploit depth cues from active gating, multi-view RGB and multi-view NIR sensing -- multi-view and gated cues across the entire spectrum. The resulting capture system consists only of low-cost CMOS sensors and flood-illumination. We propose a novel stereo-depth estimation method that is capable of exploiting these multi-modal multi-view depth cues, including the active illumination that is measured by the RCCB camera when removing the IR-cut filter. The proposed method achieves accurate depth at long ranges, outperforming the next best existing method by 39\% for ranges of 100 to \unit[220]{m} in MAE on accumulated LiDAR ground-truth. Our code, models and datasets are available \href{ https://light.princeton.edu/gatedrccbstereo/}{here}\footnote{\tiny\url{ https://light.princeton.edu/gatedrccbstereo/}\label{link}}.
\vspace{-0.8cm}
\end{abstract}

\section{Introduction}
\vspace{-1mm}
Depth estimation has become a cornerstone sensing modality for 3D scene understanding in a wide range of applications such as perception and planning in autonomous driving and robotics \cite{PseudoLidar, hu2023planning, li2023mseg3d}. Today's fully-autonomous robots mainly rely on scanning LiDAR for depth estimation~\cite{schwarz2010lidar,WaymoDataset}. However, at ranges greater than 100 m, the spatial resolution of existing sensors, with a few points per pedestrian, is not sufficient for semantic understanding.  Furthermore, both frequency-modulated as well as time-of-flight LiDAR systems have proven to be unreliable in the presence of backscatter \cite{BenchmarkLidar}.  While innovations in LiDAR technology such as MEMS scanning mechanisms~\cite{mems:mirrors:lidar:review} and advanced photodiode systems~\cite{villa2012spad} have substantially lowered costs and enabled the development of sensors with approximately 100 to 200 scanlines, they still fall short in comparison to the vertical resolution offered by modern HDR megapixel cameras, which can exceed 10k pixels.
Wide-baseline RGB stereo depth estimation methods overcome this issue by providing depth maps at image resolution, but struggle in low-light scenes and texture-less regions. 
Recently, gated imaging~\cite{heckman1967,grauer2014active,Bijelic2018,Busck2004, Busck2005, Andersson2006} has emerged as a potential alternative sensor modality for 3D detection and depth estimation, offering the capability to overcome low LiDAR-resolution, while providing comparable accuracy \cite{gated2depth2019, gated2gated, gatedstereo}. Operating in the near-infrared spectrum, gated imaging systems combine CMOS sensors with active flash illumination and analogue gated readout. This approach is robust to low-light and adverse weather conditions~\cite{Bijelic2018}.
For depth prediction, Gated2Depth~\cite{gated2depth2019} employs three gated slices in a neural network which is trained via a combination of simulation and LiDAR supervision. Following this, Walia et al.~\cite{gated2gated} proposed a self-supervised training approach resulting in higher-quality depth maps. Walz et al.~\cite{gatedstereo} recently introduced Gated Stereo, employing a wide-baseline stereo-gated configuration for depth estimation. These methods outperform scanning LiDAR systems in depth resolution, precision, and robustness to backscatter in fog, rain and snow.
While these methods successfully outperform LiDAR in depth sensing, they are constrained by the gated imager's megapixel resolution and lack of color information. This results in diminished details, particularly noticeable at long distances.
 RGB-only depth methods yield high-resolution depth maps, but these are not metric and lack the precision of LiDAR-based depth measurements.

In this work, we close this gap by proposing a low-cost CMOS-only sensing method that combines multi-view RGB sensing with gated cameras, exploiting active and multi-view cues across the visible and NIR spectrum. Specifically, we propose a NIR gated camera in conjunction with a HDR RCCB camera without an IR-cut filter present. RCCB cameras incorporate clear channel filters where conventional RGGB Bayer color filters feature the green channel, which enhances their sensitivity in low-light conditions. This joint approach allows us to use the spectral overlap for estimating high-resolution depth maps at RCCB-camera resolution of 8 megapixels, an order of magnitude higher than the gated imager resolution. Previous works have recognized the capabilities of cross-spectral imaging due to the complementary information coming from different sensor modalities \cite{bourlai2012study, he2017learning, rufenacht2013automatic, brown2011multi}. For depth estimation, however, combining images from different spectra has proven to be difficult due to the differing appearance of the images \cite{tosiRGBMultispectralMatchingDataset2022a, zhiDeepMaterialAwareCrossSpectral2018, liangUnsupervisedCrossspectralStereo2019}. Our approach combines two multi-view stereo views across the spectrum and an active illuminator (visible by both) by fusing the features of both modalities of the respective viewpoints. Specifically, to recover depth, we propose a stereo depth estimation method that incorporates a novel cross-spectral fusion module which leverages intermediate depth outputs for accurate registration of feature maps from both modalities, a pose refinement step and attention-based feature fusion. The merged features encompass complementary data from both spectra, enabling their use in the stereo network to generate accurate depth maps in any lighting conditions.

We validate our method on automotive driving data in urban, suburban and highway environments in varying illumination, and we find that the method compares favorably to existing active and hybrid methods. We also demonstrate that the high-resolution depth enables new applications, such as detecting small lost cargo objects in high-way scenarios that cannot be resolved by conventional methods.

Specifically, we make the following contributions:\vspace{-0.2eM}
\begin{itemize}
\itemsep-0.3em
  \item We propose a novel cross-spectral depth estimation approach that recovers high-resolution dense depth maps from multi-view and time-of-flight depth cues across the visible and NIR spectrum.
  \item We introduce a novel cross-modal stereo network that jointly estimates the depth from passive and active RCCB and gated features and a semi-supervised training scheme to train the estimator. 
  \item We validate that the method produces accurate depth maps on accumulated LiDAR point-clouds up to \unit[220]{m}, outperforming existing methods by 39\% in MAE for long ranges $\geq$ \unit[100]{m}. We show that these high-resolution depth estimates enable new applications such as lost cargo detection.
\end{itemize}

\section{Related Work}

\PAR{Depth Estimation from Monocular and Stereo Intensity Images.}
Depth estimation from intensity images has been thoroughly investigated using various modalities, from single-image captures \cite{MovingPeopleMovingCameras, godard2017unsupervised,li2022depthformer,guizilini20203d} to stereo images \cite{chang2018pyramid,TransformerStereo,badki2020Bi3D,yang2019hsm} and cross modal representations using intensity images augmented with sparse LiDAR data \cite{VolPropagationNetStereoLidar, SLFNetStereoLidar}. Further refinement techniques were introduced, enhancing the predicted depth maps and increasing resolution \cite{qiaoDepthSuperResolutionExplicit2023,zhaoSphericalSpaceFeature2023,aleotti2021neural,miangoleh2021boosting}.
Existing work has investigated various loss formations~\cite{godard2017unsupervised,vijayanarasimhan2017sfm, yin2018geonet, ranjan2019competitive, godard2019digging, luo2019every, dai2020self, guizilini20203d,lipson2021raft}, neural architectures \cite{Garg2016,godard2019digging,guizilini20203d,TransformerStereo,badki2020Bi3D,yang2019hsm,li2022depthformer} and introduced consistencies \cite{Garg2016, godard2017unsupervised}.
To exploit large unlabeled datasets, self-supervised approaches \cite{Zhou2017, godard2019digging,guizilini20203d, Garg2016, godard2017unsupervised} exploiting stereo- \cite{Garg2016, godard2017unsupervised} and temporal-consistencies \cite{Zhou2017, godard2019digging,guizilini20203d}.
Unfortunately, these methods do not resolve the need for dense depth ground-truth for high-quality depth estimation~\cite{chang2018pyramid,li2022depthformer, eigen2014depth,chang2018pyramid,jaritz2018sparse,ma2018sparse,MovingPeopleMovingCameras,Mayer2016,Kendall2017}. To this end, existing methods rely on sparse LiDAR measurements as ground-truth. However, using LiDAR measurements as direct inputs \cite{tang2019sparse2dense, wong2021unsupervised, park2020nonRGBLidar, hu2020PENetRGBLidar, guidenetRGBLidar, VolPropagationNetStereoLidar, SLFNetStereoLidar} for both supervised training and inference can result propagating temporal LiDAR distortions and scan pattern artifacts.

\PAR{Depth from Time-of-Flight.}
Unlike depth estimation from intensity images, Time-of-Flight (ToF) sensors determine depth by measuring the time it takes for emitted light to return to the detector. Acquisition approaches can be classified into correlation ToF cameras \cite{hansard2012time, kolb2010time, lange00tof}, pulsed ToF sensors~\cite{schwarz2010lidar}, and gated illumination with wide depth measurement bins~\cite{heckman1967,grauer2014active}.
Correlation ToF cameras use flood illumination to gauge depth from the phase difference between emitted and received light pulses, offering high spatial depth resolution \cite{hansard2012time, kolb2010time, lange00tof}. However, these sensing modalities struggle in outdoor environments due to sensitivity to ambient light.
Pulsed ToF sensors measure the round-trip time of a single light pulse to a scene point, yielding high-depth accuracy \cite{schwarz2010lidar}, however, rely on scanning that compromises spatial resolution. Moreover, these sensors degrade in fog or snow because of backscatter \cite{BenchmarkLidar,LIBRE,Jokela}.
Gated cameras combine high resolution CMOS imagers with microsecond exposure times, integrating pulsed flood-illumination with adjustable delays. Through this temporal gating, backscatter is effectively reduced \cite{Bijelic2018}, and coarse depth is reconstructed \cite{Busck2004, Busck2005, Andersson2006}. Extracting more refined depth initially focused on analytical methods~\cite{Laurenzis2007, Laurenzis2009, Xinwei2013}, Bayesian methods ~\cite{adam2017bayesian, schober2017dynamic} and deep neural networks~\cite{gated2depth2019, gated2gated} excel in low-light and outdoor scenarios. Gruber et al.\cite{gated2depth2019} predict depth using a reconstruction network rivaling conventional stereo models, while Walia et al.\cite{gated2gated} proposed a refined self-supervised method. Later, Walz et al.~\cite{gatedstereo} combined two gated imagers, optimizing depth estimation through multi-view cues. All of these methods are designed for gated imagers only, compromising resolution compared to RGB imagers and in scenarios when the NIR laser power is low compared to ambient light. We lift this limitation by combining NIR gated cameras with high-resolution visible-spectrum RCCB sensors. 

\PAR{Cross-Spectral Matching}
Conventional stereo matching algorithms assume match based on the brightness constancy assumption. However, using multiple sensors, operating in distinct spectral ranges, has been investigated as an additional source of information. Progress was reported in areas such as Face Recognition \cite{Li_2013_CVPR_Workshops, juefei2015nir, Lezama_2017_CVPR}, self-driving cars \cite{hwang2015multispectral, xu2017learning}, visual surveillance \cite{le2020edge20}, and smartphones \cite{Thavalengal_2015_CVPR_Workshops}.
Existing methods have proposed methods for matching features that may be visually distinct but remain semantically congruent \cite{ImprovingKinect,RobustStereo2011,Pinggera2012OnCS,MultiSpectral2014,MonochromeRGBStereo2016,TPAMODenseCrossModal2021,tosiRGBMultispectralMatchingDataset2022a,zhiDeepMaterialAwareCrossSpectral2018}.  Early work \cite{Pinggera2012OnCS} explores gradients as a robust feature for cross-modal matching, while \cite{MonochromeRGBStereo2016} focuses on the alignment of monochrome images, which have increased light sensitivity, with RGB images to achieve precise depth in dim lighting scenarios.
Recent methods~\cite{zhiDeepMaterialAwareCrossSpectral2018,TPAMODenseCrossModal2021, liangUnsupervisedCrossspectralStereo2019, waltersThereBackAgain2021, tosiRGBMultispectralMatchingDataset2022a} aim to learn cross-modal matching, where some aim to morph one modality directly into another~\cite{zhiDeepMaterialAwareCrossSpectral2018, waltersThereBackAgain2021}, while others propose novel descriptors for modality matching~\cite{TPAMODenseCrossModal2021,tosiRGBMultispectralMatchingDataset2022a, zhiDeepMaterialAwareCrossSpectral2018}.

\begin{figure}[t!]
    \centering
    \includegraphics[width=0.49\textwidth]{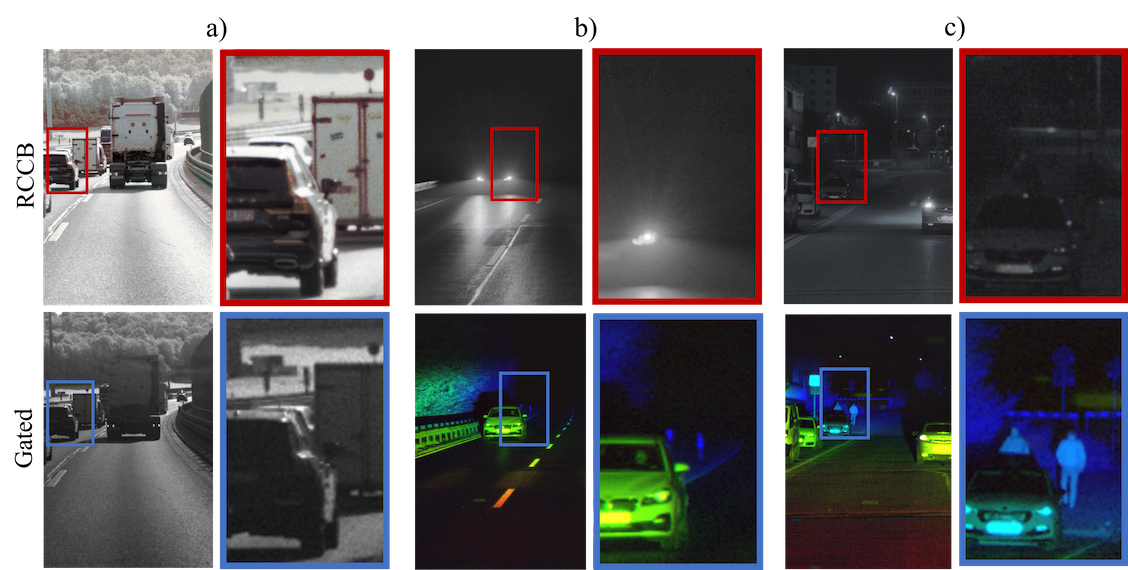}
    \caption{
     RCCB cameras (top row) capture 8 Mpix passive RGB images. Gated cameras (bottom row) record Time-of-Flight data of a scene by combining active flash illumination and analog gated readout. Both sensors are complementary, with distinct strengths depending on the scenario. RCCB cameras excel in daylight (a) with high dynamic range, resolution and color.
     At night (b, c), gated images (gated slices here RGB-color coded by mapping each slice to one RGB color) provide strong depth cues and maintain consistent scene illumination through active illumination.
     This work integrates both modalities to estimate depth accurately in all ambient illumination conditions.    
    }     \label{fig:compare}
    \vspace{-5mm}
\end{figure}

\begin{figure*}[!ht]
\vspace{-6mm}
     \centering
    \includegraphics[width=0.9\textwidth]{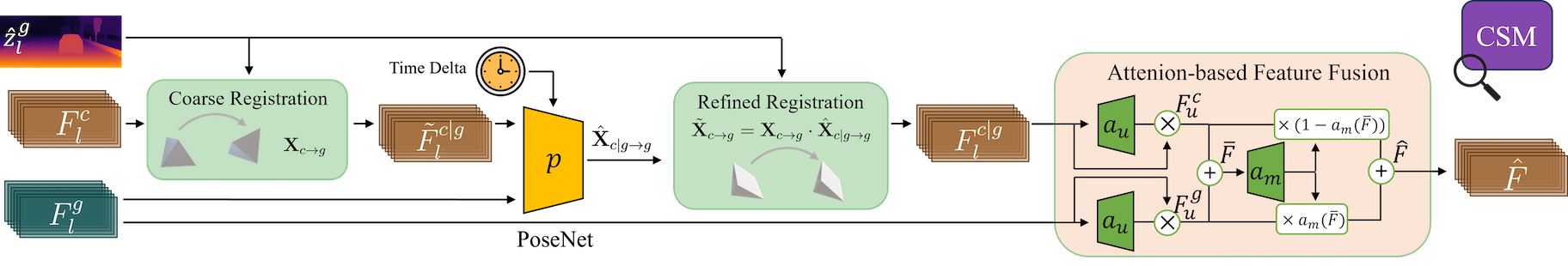}
    \vspace{-0.5eM}
    \caption{Cross-Spectral Matching (CSM). The layer fuses encoded features from RCCB ($F^c_l$) and gated ($F^g_l$) images. In the coarse registration step, RCCB features are aligned with gated features based on calibrated poses $X_{c \to g}$. Registration is refined based on residual pose $\hat{X}_{c|g \to g}$ estimated from coarse aligned images and measured time delta with PoseNet. Registered images are fused with attention-based fusion retaining complementary information in $\hat{F}$. \vspace{-5mm}}
    \label{fig:fcm}
    
\end{figure*}
\vspace{-1.0mm}

\section{Multi-view Gated and RCCB Imaging}\label{sec:GatedStereoImaging}

\vspace{-0.25eM}
We propose to image a scene with a gated camera stereo system and an RCCB stereo array characterized both by a baseline of $b=0.76$ m. 
The gated imager is an active sensor and emits a pulse of light with a confined wavelength around 808 nm, whereas the RCCB camera is a passive sensor with a sensitivity spectrum spanning the visible band from 380 - 1050 nm. 
While conventional RGB cameras use color filter arrays with an RGGB pattern, often referred to as a Bayer pattern, in RCCB cameras the green channels are replaced with clear channels. The inclusion of clear channels in this pattern allows an enhanced light sensitivity, boosting its performance $\approx$ 30\% during night-time conditions. In addition, the used Onsemi AR0820AT image sensor is optimized for both low light and challenging high dynamic range scene performance, with a 2.1 µm DR Pix BSI pixel and on-sensor 140 dB HDR capture capability.

In the stereo gated camera system, a laser pulse \( p \) is emitted at \( t = 0 \). Following a set time delay \( \xi \), the reflected scene is then integrated on both camera sensors. 
Only photons within a specific temporal gate are captured, using the gate function \(g\), embedding depth data into 2D imagery. As detailed by Gruber \textit{et al.}~\cite{gruber2018learning}, these intensities, or range-intensity-profiles \(C_k(z)\), are scene-independent and can be expressed as
\begin{equation}
\begin{aligned}
I^{k}(z,t)\;&=\;\alpha\,C_{k}(z,t),\\
    \;&=\;\alpha \int\limits_{-\infty}^{\infty} g_k(t-\xi)p_k\left(t\,-\,\cfrac{2z}{c}\right)\beta(z)dt,
\label{eq:gated_img}
\end{aligned}
\end{equation}
where \(I^{k}(z,t)\) is the gated exposure at distance \(z\) and time \(t\); \(\alpha\) represents surface reflectance, while \(\beta\) accounts for atmospheric attenuation. Both image sets are calibrated and rectified for aligned epipolar lines, enabling disparity \(d\) estimation. This disparity corresponds to distance \(z=\frac{bf}{d}\), offering depth insights across all slices.
Ambient light sources, such as sunlight or vehicle headlights influence the gated system's operation. These photons get modulated by a constant term \( \Lambda \). 
Separately, irrespective of ambient light, there is a dark current, \( D^k_v \), which is dependent on the gating settings. In total we model an image with
\begin{equation}
I^{k}_v(z)\;=\;\alpha\,C_{k}(z)+\,\Lambda + D^k_v.
\label{eq:final_gated_eq_dark_current}
\end{equation}
We follow \cite{gatedstereo}, capturing additional passive HDR images with fixed exposure times of 21 µs and 108 µs during the day, and extending these to 805 µs and 1745 µs at night.

When integrating both gated and RCCB stereo systems, each camera is represented by its calibration matrix \( K \). 
The relative orientation and position between cameras in a stereo pair are captured by the rotation matrix, \( R \in SO(3)\), and the translation vector, \( t \in \mathbb{R}^{3\times 1}\).

\section{Depth from RCCB and Gated Stereo}
\vspace{-1mm}
\noindent
In this section, we introduce our cross-modal fusion technique for depth prediction, which relies on multi-view cues from RCCB stereo and gated stereo images. By registering and fusing cross-spectral features through an attention mechanism and prior pose refinement within the stereo network, we capitalize complementary information from different camera modalities in Sec. \ref{sec:CrossSpectralMatching}. We integrate this feature fusion in a stereo network described in Sec.~\ref{sec:Network} which we jointly train uni-modal and multi-modal, facilitating a holistic feature representation across modalities and minimizing domain differences between modalities as detailed in Sec.~\ref{sec:DomainGap}. The training approach is detailed in Sec.~\ref{sec:consistencylosses}.  

\subsection{Cross-Spectral Matching} \label{sec:CrossSpectralMatching}
\noindent
We align and combine cross-modal features in a two-stage approach, where we warp features first into a shared space based on a refined pose. With these aligned features in hand, we perform an attention-based fusion as input to the remainder of the stereo estimation network. An overview of cross-spectral matching (CSM) is illustrated in Fig.~\ref{fig:fcm}.

\PAR{Feature Extraction and Alignment.}\label{sec:DomainGap}
We utilize two feature extractor backbones for color $f_b^c$ and gated $f_b^g$, and share the weights for each view $I^m_l, I^m_r$ for $m \in\ \lbrace c,g \rbrace$, that is
\begin{align}
    f_b^c&: I^c_l, I^c_r \to F^c_l, F^c_r, \label{eq:ref_feat1}\\
    f_b^g&: I^g_l, I^g_r \to F^g_l, F^g_r. \label{eq:ref_feat2}
\end{align}
As a feature extractor, we use MPViT \cite{lee2022mpvit}, a powerful vision transformer for dense prediction tasks.
To align the features, we use the pose information from camera calibration \( \mathbf{X}_{x\to g} \) and an intermediate depth estimation \( \hat{\mathbf{z}}^g_l \) from an iterative depth estimation method, see Section~\ref{sec:Network}, to warp corresponding views. The mapping for homogeneous coordinates $x_g$ and $x_c$ from $I^g_l$ and $I^c_l$ is defined as 
\begin{equation}
    x_g \sim K_c \textbf{X}_{c \to g} \hat{\textbf{z}}^g_l K_g^{-1} x_c,
\end{equation}
where $K_c$ and $K_g$ are the camera matrices of the gated and RCCB camera, and 
\(
\textbf{X}_{c \to g} = 
\begin{psmallmatrix}
R_{c \to g} & t_{c \to g}\\
0 & 1
\end{psmallmatrix}\) with \(R_{c \to g} \in SO(3)\) and \(t_{c \to g} \in \mathbb{R}^{3\times 1}\).
We transform the features of the left RCCB camera, denoted $F^c_l$, to match the features of the left gated camera $F^g_l$, thus creating $\tilde{F}^{{c|g}}_l$.

\PAR{Pose Refinement.} The RCCB stereo camera and the gated stereo camera are independently synchronized to microsecond precision. However, the RCCB camera may accumulate a slight offset of up to 20 milliseconds between images because of automatic exposure and shutter timing. To address this misalignment, we utilize a lightweight Convolutional Neural Network (CNN) framework dubbed PoseNet $p$. This framework estimates the rotational and translational adjustments necessary for alignment, based on prealigned feature maps.
The input to $p$ is the concatenated context $F^g_l$, the transformed $\tilde{F}^{{c|g}}_l$ and the measured time offset $t$ between modalities. The time  is integrated into every downsampled layer of the pose network as additional channel, except for the final layer. This channel uniformly replicates the value of the time-offset across the spatial dimension. The computed pose update, denoted as $\hat{\textbf{X}}_{c|g \to g}$ combines the initial pose as
\(\tilde{\textbf{X}}_{c \to g} = \textbf{X}_{c \to g} \cdot \hat{\textbf{X}}_{c|g \to g}\). Subsequently, a second warping operation with the mapping 
\begin{equation}
    x_g \sim K_c \tilde{\textbf{X}}_{c \to g} \hat{\textbf{z}}^g_l K_g^{-1} x_c,
\end{equation}
is applied, which generates the aligned features $F^{{c|g}}_l$.

\begin{figure*}[!ht]
\vspace{0mm}
   \centering
    \begin{center}
        \includegraphics[width=\textwidth]{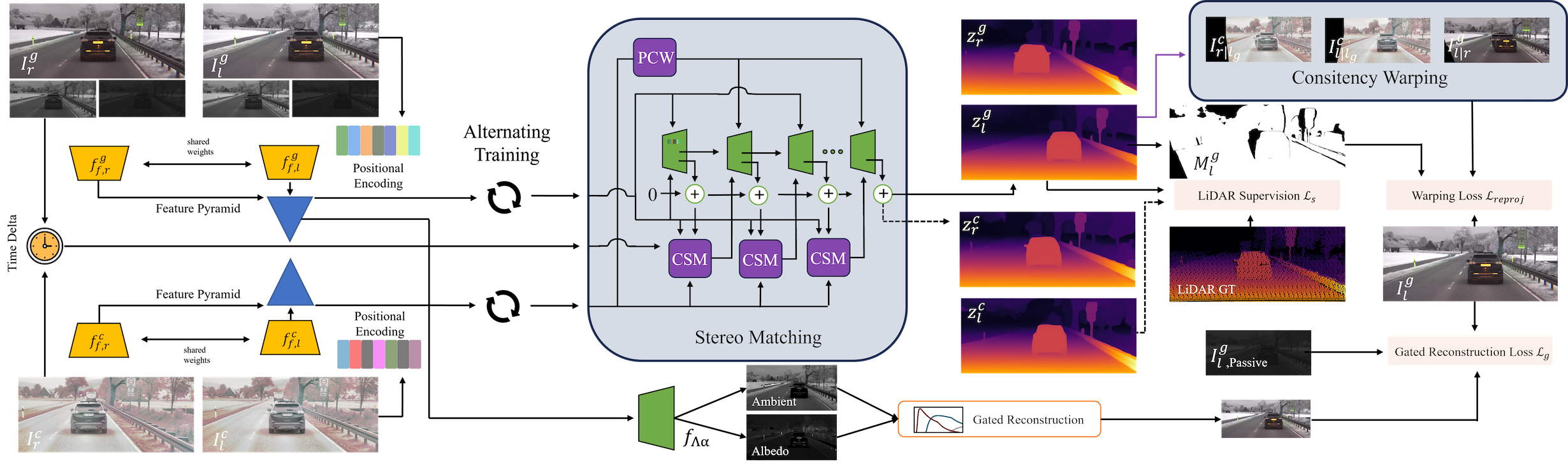}
    \end{center}
    \vspace{-2eM}
    \caption{The proposed cross-spectral stereo architecture for depth estimation from stereo RCCB and stereo gated images incorporating our CSM layer. The network can output depth for all four input images. Intermediate depth estimates are used for iterative fusion within the CSM along the depth estimation process. The network is trained with self-supervision (Left-Right consistency for RCCB and gated images, Gated Reconstruction) and LiDAR supervision.}
    \label{fig:arch}
    \vspace{-5mm}
    
\end{figure*}

\PAR{Attention-based Feature Fusion.} 
Following the alignment, we fuse RCCB and gated features, aiming to combine contextual information from both spectra effectively. Our approach adopts a two-step process. Firstly, we employ channel self-attention for aggregating both global and local contexts within feature maps.
Secondly, we combine the individual feature maps, utilizing the predicted attention weights. 
The first setup is defined as
\begin{align}
    \bar{F} &= \frac{F_u^g \oplus F_u^c}{a_u(F^g_l) + a_u(F^c_l)} \\
    F_u^g &= F^g_l \otimes a_u(F^g_l) \\
    F_u^c &= F^{c|g}_l \otimes a_u(F^{c|g}_l),
\end{align}
where $\oplus$ denotes element-wise addition and $\otimes$ indicates element-wise multiplication, and the attention $a_u()$ is calculated following \cite{dai2021attentional}. 
The final fusion of features $\hat{F}$ are the result of the following weighting-operation
\begin{equation}
    \hat{F} = \left(F_u^g \otimes a_m(\bar{F})\right) \oplus \left(F_u^c \otimes (1 - a_m(\bar{F}))\right),
\end{equation}
where $a_m$ follows the implementation as in \cite{dai2021attentional} and denotes the multi-modal attention network, facilitating the effective combination of features from both modalities.

\subsection{Stereo Matching}\label{sec:Network}
With the process to align features $\hat{F}$ in hand, we predict depth across all camera views. This task is executed through a stereo matching network, as depicted in Figure~\ref{fig:arch}. We build on top of the CREStereo architecture \cite{liPracticalStereoMatching2022} with major modifications to allow the development of a dynamic framework switching between modalities. This dynamic interchangeability allows us to adapt and optimize the disparity prediction in either modalities coordinate system. Such flexibility not only enhances domain generalization but also opens avenues for the application of various consistency losses, thereby improving the accuracy of our predictions.

To bridge the coordinate system we heavily rely on the CSM layers, whose predicted context feature maps are used to guide the prediction in the targeted frames. 
Thereby, we rely on the iterative refinement introduced in \cite{liPracticalStereoMatching2022} and calculate the correlation volume in each step according to \cite{liPracticalStereoMatching2022}. Here, we predict the correlation in the adaptive group correlation layers uni-modal and alternate in modality through the iterative refinement. The secondary modality is projected into the target frame with the refined transformation $\tilde{\textbf{X}}_{c \to g}$ in the pre-correlation warping PCW, see Fig.~\ref{fig:arch}.

Then the correlation is calculated as follows,
\begin{equation}
    \vspace{-1mm}
    \text{Corr}(x,y,k) = \frac{1}{C}\sum_{i=1}^{C} F^{v}_l(i, x, y) F^{v}_r (i, x', y'),
\end{equation}
where $F^{v}$ is the respective feature map transformed into the modality $v\in\lbrace c,g,c|g,g|c\rbrace$, with camera view $l,r$. We follow 
\cite{liPracticalStereoMatching2022} and predict $x'=x+f(k)$, $y'=y+g(k)$, with fixed offsets $f(k)$ and $g(k)$ for the $k$-th correlation pair, sum all channels $C$ and apply 2D-1D alternate local search strategy for computational efficiency. 
Notably, the initial iteration at the coarsest scale focuses on predicting depth solely from the target modality.

\subsection{Training Supervision}
\label{sec:consistencylosses}
\noindent
The network is trained to output the disparity $d$ which is converted into the depth $z$ for all modalities $g,c$ and views $l,r$.
In addition we train the stereo matching uni-modal and multi-modal, with and without cross-spectral feature enhancement to ensure optimal extracted features while sharing the stereo matching stage. This is achieved by deactivating the CSM and PCW layers. 
Through this alternating training, we ensure  that the backbone learns relevant features for all modalities and the mix and matching between modalities forces all features to be domain independent, thereby creating robust cross-modal representations. 
Further this allows us to implement self-supervised and supervised loss functions for both the gated camera and the RCCB camera, as well as consistencies in between. 

All self-supervised consistency losses and supervised losses are described below. Without diminishing generality, in the following all losses are defined for disparity prediction in the gated frame for better readability. 

\PAR{Left-Right Reprojection Consistency.}
The projection loss enforces the photometric consistency between the left and right camera views within each modality. Cross-modally the homogeneity between predicted depth maps is enforced. The total loss for the left gated camera $g_l$ can be written as,
\begin{equation}
\begin{aligned}
\mathcal{L}_{w}^{g_l} &= \mathcal{L}_p(I_l^g, I_{r|l_g}^g) + \mathcal{L}_p(I_{l|l_g}^c, I_{r|l_g}^c) + \mathcal{L}_p(z^{c|g}_l, z_l^g),
\label{eq:reproj_loss}
\end{aligned}
\end{equation}
with $I_{r|l_g}^g$ the $r$ right $g$ gated image warped into the $l$ left gated view using the predicted depth ${z}_{l}^g$ denoted as warping operation $_{l_g}$ for the stereo pairs. For the gated warping consistency further the RCCB frames $I^C$ are warped according to the predicted depth into the gated frame $_{l_g}$. Additionally, the predicted depth in $c$ is transformed to the gated frame $g$ leading to $z^{c|g}_l$. 
Consistencies are also applicable to the right gated frame, yielding \(\mathcal{L}^{g_r}_w\), and to both left \(\mathcal{L}^{c_l}_w\) and right \(\mathcal{L}^{c_r}_w\) RCCB frames. The total loss can be written as $\mathcal{L}_{reproj}=\mathcal{L}^{c_l}_w +\mathcal{L}^{c_r}_w+\mathcal{L}^{g_l}_w +\mathcal{L}^{g_r}_w$.
Note, $\mathcal{L}_p$ follows \cite{godard2017unsupervised} and is a similarity loss based on the structural similarity (SSIM) metric \cite{ssim} and the $L_1$ norm, $\mathcal{L}_p(a,b)=0.85\frac{1 - SSIM(a, b)}{2} + 0.15\|a - b\|_1$.

\PAR{Gated Reconstruction Loss.}
To supervise the embedded time of flight information in the gated slices we adopt the cyclic gated reconstruction loss from~\cite{gated2gated}, which uses measured range intensity profiles to reconstruct the input gated images from the predicted depth $z$, the albedo $\tilde{\alpha}$, and the ambient $\tilde{\Lambda}$. Departing from~\cite{gated2gated} who employed measured profiles, we employ an analytical gating model. We estimate the albedo $\tilde{\alpha}$ and ambient $\tilde{\Lambda}$ through an additional context encoder taking the feature pyramid as input, see Figure~\ref{fig:arch}, and model a gated slice as \(\tilde{I}^k(z)\; =\;\tilde{\alpha}\,C_k(z) + \tilde{\Lambda}.\)
The loss term incorporates both per-pixel difference and structural similarity, following
\begin{equation}
\begin{aligned}
\mathcal{L}_{recon} = \mathcal{L}_{p}(M_{g} \odot \tilde{I}^k(z), M_{g} \odot I^k)+\mathcal{L}_p(\tilde{\Lambda},\Lambda^{k_0}),
\label{eq:gated_reconstruction_loss}
\end{aligned}
\end{equation}
with $M_g$ as per-pixel SNR consistency mask~\cite{gated2gated}.

\PAR{LiDAR Supervision.} 
We supervise final and intermediate disparity predictions. Each disparity prediction $\{\mathbf{d}_{l,i},...,\mathbf{d}_{l,n} \}$ is upsampled to full resolution and compared to ground-truth with a weighted combination of $l_1$ and $l_2$ defined as,
\begin{equation}
    \mathcal{L}_{lidar} = \sum_{i=1}^{n} \gamma^{n-i} \Big(\frac{2}{3}||{d}_{gt}-M\odot {d}_l||_1  
    + \frac{1}{3}||{d}_{gt}-M\odot {d}_l||_2^2 \Big).
\end{equation}
The weight $\gamma$ is set to 0.9 and the mask $M$ excludes areas without ground-truth. For ground-truth we use accumulated and sparse LiDAR measurements, more details in Section~\ref{sec:dataset}.

\PAR{Overall Training Loss.}
The following loss term is obtained by combining all self-supervised and supervised loss components from above,
\begin{align}
\mathcal{L}_{stereo} = c_1 \mathcal{L}_{reproj} +  c_2 \mathcal{L}_{recon} + c_3 \mathcal{L}_{lidar},
\label{eq:total_stereo_loss}
\end{align}
which we combine with scalar weights $c_{1, \ldots, 3}$ provided in the Supplemental Material. 

\PAR{Implementation Details.}
We refer to the Supplemental Document for implementation details, training settings, and hyperparameter settings used for the approach described.

\begin{table}[!t]
    \vspace{-0.5eM}
    \footnotesize
    \setlength{\tabcolsep}{4pt} 
    \setlength\extrarowheight{2pt}
    \centering
    \resizebox{.99\linewidth}{!}{
    \begin{tabular}{@{}c|lcccccccc@{}}
            \toprule
            & \multirow{2}{*}{\textbf{\textsc{Method}}} & \textbf{Modality} & \textbf{Train} & \textbf{RMSE}     & \textbf{ARD}   & \textbf{MAE}  & $\boldsymbol{\delta_1}$ & $\boldsymbol{\delta_2}$ & $\boldsymbol{\delta_3}$  \\ 
			&& &  & $\left[ m \right]$  &  & $\left[ m \right]$ & $\left[ \% \right]$ & $\left[ \% \right]$ & $\left[ \% \right]$\\
			\midrule
			\multicolumn{10}{c}{\textbf{Test Data -- Night (Evaluated on LiDAR Ground-Truth Points)}} \\
			\midrule
			\multirow{18}{*}{\rotatebox[origin=l]{90}{\parbox[c]{6.5cm}{\centering \textbf{\textsc{Comparison to state-of-the-art}}}}} 
			 & \textsc{Gated2Depth} \cite{gated2depth2019} & Mono-Gated & D & 16.15  &  0.17  &  8.07  &  75.70  &  92.74  &  96.47  \\ 
 & \textsc{Gated2Gated}  \cite{gated2gated} & Mono-Gated & MG & 14.08  &  0.19  &  7.95  &  79.84  &  92.95  &  96.59  \\ 
 & \textsc{PackNet} \cite{guizilini20203d} & Mono-RGB & M & 17.82  &  0.20  &  10.21  &  66.35  &  87.85  &  95.61  \\ 
 & \textsc{Monodepth2} \cite{godard2019digging} & Mono-RGB & M & 18.44  &  0.18  &  9.47  &  75.70  &  90.46  &  95.68  \\ 
 & \textsc{SimIPU} \cite{li2022simipu} & Mono-RGB & D & 15.78  &  0.18  &  8.71  &  76.25  &  90.84  &  96.44  \\ 
 & \textsc{AdaBins} \cite{bhat2021adabins} & Mono-RGB & D & 14.45  &  0.15  &  7.58  &  81.47  &  93.75  &  97.39  \\ 
 & \textsc{DPT} \cite{ranftl2021vision} & Mono-RGB & D & 12.15  &  0.12  &  6.31  &  85.38  &  95.94  &  98.42  \\ 
 & \textsc{DepthFormer} \cite{li2022depthformer} & Mono-RGB & D & 12.15  &  0.11  &  6.20  &  85.18  &  95.76  &  98.47  \\ 
 & \textsc{PSMNet} \cite{Chang2018} & Stereo-RGB & D & 27.98  &  0.27  &  16.02  &  50.77  &  74.77  &  85.93  \\ 
 & \textsc{STTR} \cite{li2021revisiting} & Stereo-RGB & D & 20.99  &  0.19  &  11.14  &  70.84  &  87.70  &  93.46  \\ 
 & \textsc{HSMNet} \cite{yang2019hsm} & Stereo-RGB & D & 12.42  &  0.09  &  5.87  &  88.41  &  96.08  &  98.50  \\ 
 & \textsc{ACVNet} \cite{xu2022attention} & Stereo-RGB & D & 11.70  & { 0.08 } &  5.25  &  89.91  &  96.33  &  98.47  \\ 
 & \textsc{RAFT-Stereo} \cite{lipson2021raft} & Stereo-RGB & D & 10.89  &  0.09  & 5.10 & 90.47 &  96.71  &  98.64  \\ 
 & \textsc{CS-Stereo} \cite{zhiDeepMaterialAwareCrossSpectral2018} & RCCB-NIR & D & 21.35 & 0.20 & 11.48 &  72.73  &  89.71  &  95.58  \\ 
 & \textsc{UCSSM} \cite{liangUnsupervisedCrossspectralStereo2019} & RCCB-NIR & D & 18.22 & 0.27 & 14.63 &  64.51  &  87.12  &  94.27  \\ 
  & \textsc{CREStereo} \cite{liPracticalStereoMatching2022} & Stereo-RCCB & D & 12.05 & 0.10 & 5.18 & 88.48 & 94.12 & 97.26  \\ 
 & \textsc{Gated Stereo \cite{gatedstereo}} & Stereo-Gated & DGS & \underline{6.39 } & \underline{ 0.05 } & \underline{ 2.25 } & \underline{ 96.40 } &  \underline{98.44} & \underline{99.24} \\
 & \textbf{\textsc{Gated RCCB Stereo}} & Stereo-RCCB-Gated & DGS & \textbf{6.23} & \textbf{0.04} & \textbf{2.03} & \textbf{96.69} & \textbf{98.50} & \textbf{99.26}

\\
			\midrule
			\multicolumn{10}{c}{\textbf{Test Data -- Day (Evaluated on LiDAR Ground-Truth Points)}} \\
			\midrule
			\multirow{18}{*}{\rotatebox[origin=l]{90}{\parbox[c]{6.5cm}{\centering \textbf{\textsc{Comparison to state-of-the-art}}}}}
             & \textsc{Gated2Depth} \cite{gated2depth2019} & Mono-Gated & D & 28.68  &  0.22  &  14.76  &  66.68  &  82.76  &  87.96  \\ 
 & \textsc{Gated2Gated}  \cite{gated2gated} & Mono-Gated & MG & 16.87  &  0.21  &  9.51  &  73.93  &  92.15  &  96.10  \\ 
 & \textsc{PackNet} \cite{guizilini20203d} & Mono-RGB & M & 17.69  &  0.21  &  9.77  &  72.12  &  90.65  &  96.51  \\ 
 & \textsc{Monodepth2} \cite{godard2019digging} & Mono-RGB & M & 20.78  &  0.22  &  10.06  &  79.05  &  90.66  &  94.69  \\ 
 & \textsc{SimIPU} \cite{li2022simipu} & Mono-RGB & D & 14.33  &  0.14  &  7.50  &  81.77  &  94.01  &  97.92  \\ 
 & \textsc{AdaBins} \cite{bhat2021adabins} & Mono-RGB & D & 12.76  &  0.12  &  6.53  &  86.15  &  95.77  &  98.41  \\ 
 & \textsc{DPT} \cite{ranftl2021vision} & Mono-RGB & D & 11.29  &  0.09  &  5.52  &  89.56  &  96.83  &  98.79  \\ 
 & \textsc{DepthFormer} \cite{li2022depthformer} & Mono-RGB & D & 10.59  &  0.09  &  5.06  &  90.65  &  97.46  &  99.02  \\ 
 & \textsc{PSMNet} \cite{Chang2018} & Stereo-RGB & D & 32.13  &  0.28  &  18.09  &  53.82  &  74.91  &  84.96  \\ 
 & \textsc{STTR} \cite{li2021revisiting} & Stereo-RGB & D & 16.77  &  0.16  &  8.99  &  78.44  &  93.53  &  98.01  \\ 
 & \textsc{HSMNet} \cite{yang2019hsm} & Stereo-RGB & D & 10.36  &  0.08  &  4.69  &  92.47  &  97.93  &  99.11  \\ 
 & \textsc{ACVNet} \cite{xu2022attention} & Stereo-RGB & D & 9.40  & { 0.07 } &  4.08  & { 94.61 } &  98.36  & 99.12 \\ 
 & \textsc{RAFT-Stereo} \cite{lipson2021raft} & Stereo-RGB & D & 9.40 & { 0.07 } & 4.07 &  93.76  &  98.15  &  99.09  \\
  & \textsc{CS-Stereo} \cite{zhiDeepMaterialAwareCrossSpectral2018} & RCCB-NIR & D & 21.51 & 0.22 & 11.87 &  73.70  &  88.77  &  96.06  \\ 
   & \textsc{UCSSM} \cite{liangUnsupervisedCrossspectralStereo2019} & RCCB-NIR & D & 17.32 & 0.29 & 13.26 &  64.80  &  84.78  &  93.83  \\ 
 & \textsc{CREStereo} \cite{liPracticalStereoMatching2022} & Stereo-RCCB & D & 9.68 & 0.06 & 3.88 & 95.02 & 96.04 & 98.57 \\ 
  & \textsc{Gated Stereo \cite{gatedstereo}} & Stereo-Gated & DGS & \underline{7.11 } & \underline{ 0.05 } & \underline{ 2.25 } & \underline{ 96.87 } & \underline{98.46 } &  \underline{99.11}  \\
 & \textbf{\textsc{Gated RCCB Stereo}} & Stereo-RCCB-Gated & DGS & \textbf{6.89} & \textbf{0.03} & \textbf{1.95} & \textbf{97.18} & \textbf{98.55} & \textbf{99.18}\\
			\midrule
            \bottomrule
    \end{tabular}
    }
    \vspace*{-5pt}
    \caption{\label{tab:results_g2d}\small Evaluation of the method and competing gated approaches on~\cite{gatedstereo}. We compare our model to supervised and unsupervised approaches. ``M'' refers to methods that use temporal data for training, S for stereo supervision, ``G'' for gated consistency and ``D'' for depth supervision. Best results in each
    category are in \textbf{bold} and second best are
    \underline{underlined}.
	\vspace*{-1mm}
}
\end{table}

\begin{table}[!t]
    \vspace{-0.2eM}
    \footnotesize
    \setlength{\tabcolsep}{4pt} 
    \setlength\extrarowheight{2pt}
    \centering
    \resizebox{.8\linewidth}{!}{
    \begin{tabular}{@{}c|lcc|cc|cc@{}}
            \toprule
            \multicolumn{2}{c}{\textsc{Evaluation Range}} & \multicolumn{2}{c|}{\textbf{0 - 160 m}} & \multicolumn{2}{c|}{\textbf{0 - 220 m}} & \multicolumn{2}{c}{\textbf{100 - 220 m}} \\
			\midrule
            & \multirow{1}{*}{\textbf{\textsc{Method}}} &  \textbf{RMSE}       & \textbf{MAE}  &  \textbf{RMSE}       & \textbf{MAE}  &  \textbf{RMSE}        & \textbf{MAE}  \\ 
			\midrule
			\multirow{3}{*}{\rotatebox[origin=l]{90}{\parbox[c]{1.2cm}{\centering \textbf{\textsc{night}}}}}
             & \textsc{CREStereo} \cite{liPracticalStereoMatching2022} & 13.58 & 8.60 & 17.64 & 10.05  & 26.39 & 20.24 \\
  & \textsc{Gated Stereo} \cite{gatedstereo}  & \underline{11.45} & \underline{7.36}  & \underline{14.03} & \underline{8.93}  & \underline{25.55} & \underline{18.36}  \\
 & \textbf{\textsc{Gated RCCB Stereo}} & \textbf{10.74} & \textbf{7.02} & \textbf{12.02} & \textbf{7.94}  & \textbf{15.67} & \textbf{11.15}

\\
			\midrule
			\multirow{3}{*}{\rotatebox[origin=l]{90}{\parbox[c]{1.2cm}{\centering \textbf{\textsc{day}}}}}
             & \textsc{CREStereo} \cite{liPracticalStereoMatching2022}& 11.16 & 6.53 & 15.65 & \underline{8.11} & \underline{20.76} & \underline{14.65} \\ 
& \textsc{Gated Stereo} \cite{gatedstereo} & \underline{10.75} & \underline{6.42} & \underline{14.24} & 8.67 & {22.07} &  {16.79}  \\ 
 & \textbf{\textsc{Gated RCCB Stereo}}  & \textbf{9.72} & \textbf{6.24}  & \textbf{10.69}  & \textbf{6.83} & \textbf{14.33} &  \textbf{10.07}

\\
			\midrule
            
            \bottomrule
    \end{tabular}
    }
    \vspace*{-5pt}
    \caption{\label{tab:results_g2d_acc}\small Evaluation on \emph{Accumulated LiDAR Scans}. We compare our method to the top 3 methods from Tab.~\ref{tab:results_g2d} using accumulated dense LiDAR as ground-truth for a range from 0 - 220 m.
	\vspace*{-5mm}
}
\end{table}
\begin{table*}[!t]
    \vspace*{-5mm}
	\centering
	\vspace{-0.2eM}
	\includegraphics[width=1.02\textwidth]{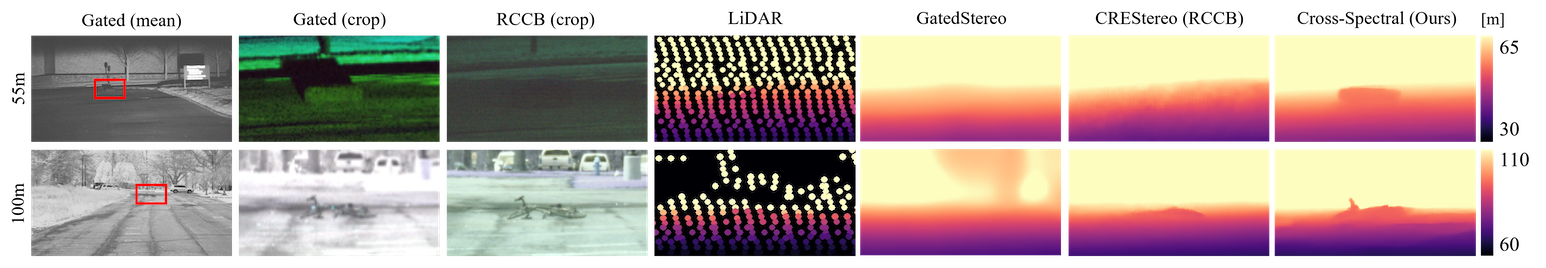}\vspace{-0.9eM}
	\captionof{figure}{Depth estimation for "lost cargo", small objects at far distances on ground level that may be lost from preceding vehicles. Our method estimates accurate depth for these small objects in both daylight and nighttime conditions by integrating complementary RCCB and gated images. Single modality methods suffer from limitations: CREStereo~\cite{liPracticalStereoMatching2022} (RCCB) lacks effective illumination at night, and Gated Stereo~\cite{gatedstereo} suffers from poor resolution during the day.}
	\label{fig:lost_cargo}
	\vspace*{-0.4cm }
\end{table*}

\begin{figure}[!t]
   \centering
   \includegraphics[width=0.82\linewidth]{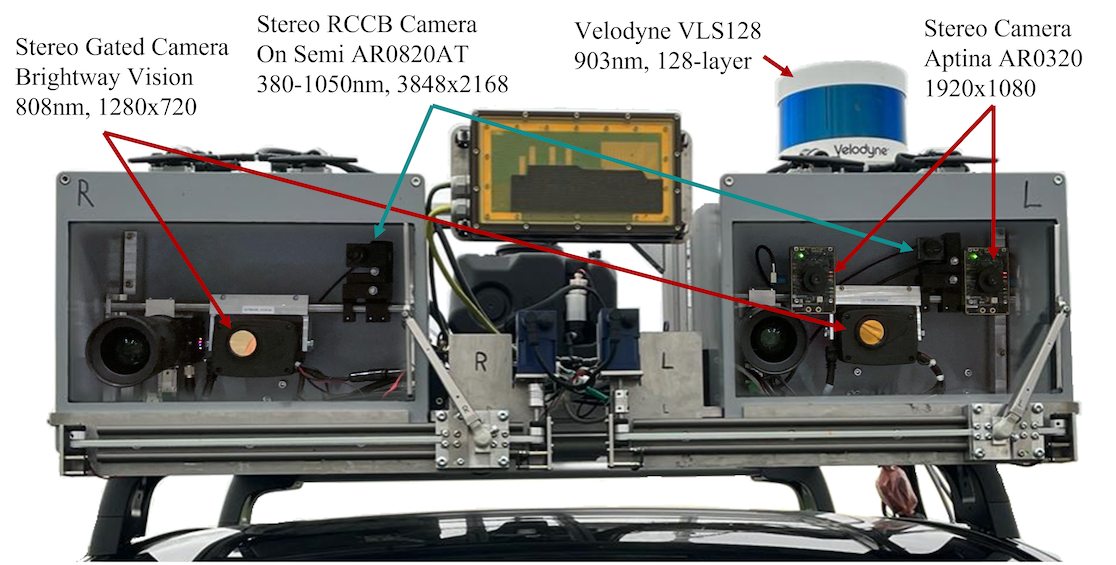}
   \vspace{-3mm}
   \caption{The sensor setup of the test vehicle used for capturing the dataset. It features a stereo gated camera, consisting of a flood-light flash source (not visible, mounted at front bumper of the car) and two gated imagers, a Velodyne VLS128 scanning lidar, a standard stereo RGB camera and the RCCB stereo camera.}
   \label{fig:test_vehicle}
   \vspace{-4mm}
\end{figure}

\vspace{-5pt}
\section{Dataset}
\label{sec:dataset}
\noindent
For training and testing, we use the dataset introduced by Walz et al. \cite{gatedstereo}. The dataset includes stereo gated, stereo RGB and ground-truth LiDAR data. More information is given in the Supplemental Material. 
In this work, we extend the dataset with RCCB stereo data, captured with an AR0820 sensor. All sensors were housed in a portable sensor cube as showcased in Figure~\ref{fig:test_vehicle}. As an additional source of ground-truth, we utilize a densely constructed LiDAR map, derived from a custom adaptation of the LIO-SAM algorithm, as detailed in Shan et al. \cite{shan2020lio}. We refer to the Supplemental Document for details on the setup and dataset.

\vspace{-2mm}

\section{Assessment}
\noindent
In this section, we experimentally validate our proposed method. We examine the accuracy of our depth estimation under nighttime and daytime conditions and compare it to existing depth estimation techniques. Furthermore, we validate our design choices through a series of ablation studies.

\PAR{Experimental Setup.}
The test set comprises 2463 frames, split into 1269 daytime and 1194 nighttime frames. Each frame is accompanied by high-resolution LiDAR ground-truth measurements, capturing reliable data up to 160 m. We further use the 655 frames of refined LiDAR ground-truth (303 daytime and 352 nighttime). These frames feature accumulated point clouds, allowing us to assess the methods on a dense ground-truth for accuracy up to a distance of \unit[220]{m}. Our method's evaluated depth maps show the perspective of the left gated camera and match the resolution of the RCCB images.

Our evaluation metrics are in line with those established in \cite{eigen2014depth}. We use Root Mean Square Error (RMSE), Mean Absolute Error (MAE), Absolute Relative Difference (ARD), and the threshold accuracy metric $\delta_i < 1.25^i$ for $i \in {1, 2, 3}$. 
All methods in our evaluation have been fine-tuned on our dataset for a fair comparison.

\begin{table}[!t]
	\centering
	\setlength{\tabcolsep}{2pt}
	\resizebox{1.01\linewidth}{!}{
		\begin{tabular}{l|lcccccccccc}
			\toprule
			& \multirow{2}{*}{\textbf{Modality}} &  \textbf{Full} &   \textbf{CS} &  \textbf{Pose} &  \textbf{Att.} & \textbf{MPViT} &  \textbf{RMSE} &  \textbf{MAE}  & $\boldsymbol{\delta_1}$ & $\boldsymbol{\delta_2}$ & $\boldsymbol{\delta_3}$  \\ 
			& & \textbf{Res.} & \textbf{Training} & \textbf{Ref.} & \textbf{Fusion} & \textbf{Backb.} & $\left[ m \right]$ & $\left[ m \right]$ & $\left[ \% \right]$ & $\left[ \% \right]$ & $\left[ \% \right]$\\
			\midrule
			\multicolumn{11}{c}{\textbf{Test Data -- Night (Evaluated on LiDAR Ground-Truth Points)}} \\
			\midrule
			\midrule
			\multirow{8}{*}{\rotatebox[origin=l]{90}{\parbox[c]{3.35cm}{\centering \textbf{\textsc{Ablation}}}}} 
			& Stereo-RCCB-Gated & \cmark & \cmark & \cmark & \cmark & \cmark & \textbf{6.23} & \textbf{2.03} & \textbf{96.69} & \textbf{98.50} & \textbf{99.26}  \\ 
& Stereo-RCCB-Gated & \xmark & \cmark & \cmark & \cmark & \cmark & \underline{6.53}  & \underline{2.04} & \underline{96.37} &  \underline{98.45} & \underline{99.24}  \\ 
& Stereo-RCCB-Gated & \xmark & \xmark & \cmark & \cmark & \cmark & 6.87 & 2.18 & 96.20 & 98.24 & 99.13  \\ 
& Stereo-RCCB-Gated & \xmark & \xmark & \xmark & \cmark & \cmark & 6.98 & 2.23 & 96.01 & 98.21 & 99.11  \\ 
& Stereo-RCCB-Gated & \xmark & \xmark & \xmark & \xmark & \cmark & 7.23  & 2.42 & 95.89 & 98.20 & 99.10  \\ 
& Stereo-RCCB-Gated & \xmark & \xmark & \xmark & \xmark & \xmark & 8.17 & 2.74 & 95.23 & 97.79 & 98.89  \\ 
& RCCB-Gated & \xmark & \xmark & \xmark & \xmark & \xmark & 10.56 & 7.89 & 45.23 & 79.49 & 91.14  \\ 
& Mono-Gated & \xmark & \xmark & \xmark & \xmark & \xmark & 10.87 & 4.70 & 89.91 & 95.77 & 97.90\\
			\midrule
			\multicolumn{11}{c}{\textbf{Test Data -- Day (Evaluated on LiDAR Ground-Truth Points)}} \\
			\midrule
			\multirow{8}{*}{\rotatebox[origin=l]{90}{\parbox[c]{3.35cm}{\centering \textbf{\textsc{Ablation}}}}}
            & Stereo-RCCB-Gated & \cmark & \cmark & \cmark & \cmark & \cmark & \textbf{6.89} & \underline{1.95} & \textbf{97.18} & \textbf{98.55} & \textbf{99.18}   \\ 
& Stereo-RCCB-Gated & \xmark & \cmark & \cmark & \cmark & \cmark & \underline{7.09} & \textbf{1.93} & \underline{97.03} & \underline{98.46} & \underline{99.11}  \\ 
& Stereo-RCCB-Gated & \xmark & \xmark & \cmark & \cmark & \cmark & 7.57 & 2.12 & 96.62 & 98.35 & 99.04  \\ 
& Stereo-RCCB-Gated & \xmark & \xmark & \xmark & \cmark & \cmark & 7.64 & 2.16 & 96.37 & 98.52 & 99.06  \\ 
& Stereo-RCCB-Gated & \xmark & \xmark & \xmark & \xmark & \cmark & 7.92 & 2.29 & 96.50 & 98.25 & 98.00 \\ 
& Stereo-RCCB-Gated & \xmark & \xmark & \xmark & \xmark & \xmark  & 8.17 & 2.44 & 96.46 & 98.12 & 98.92 \\ 
& RCCB-Gated & \xmark & \xmark & \xmark & \xmark & \xmark & 8.61 &  4.73 & 67.33 & 89.75 & 96.30  \\ 
& Mono-Gated & \xmark & \xmark & \xmark & \xmark & \xmark & 13.71 & 6.05 & 88.99 & 95.56 & 97.71\\
			\midrule
            \bottomrule
			
	\end{tabular}}
	\vspace*{-5pt}
\caption{Ablation Experiments on the dataset from~\cite{gatedstereo}. We investigate different resolution, training method, remove components of our proposed CSM, the MPViT \cite{lee2022mpvit} backbone and test different input modalities.}
	\label{tab:ablation}
	\vspace{-5mm}
\end{table}

\
\begin{figure*}[!t]
    \vspace*{-3mm}
	\centering
	\vspace{-0.2eM}
	\includegraphics[width=1\textwidth]{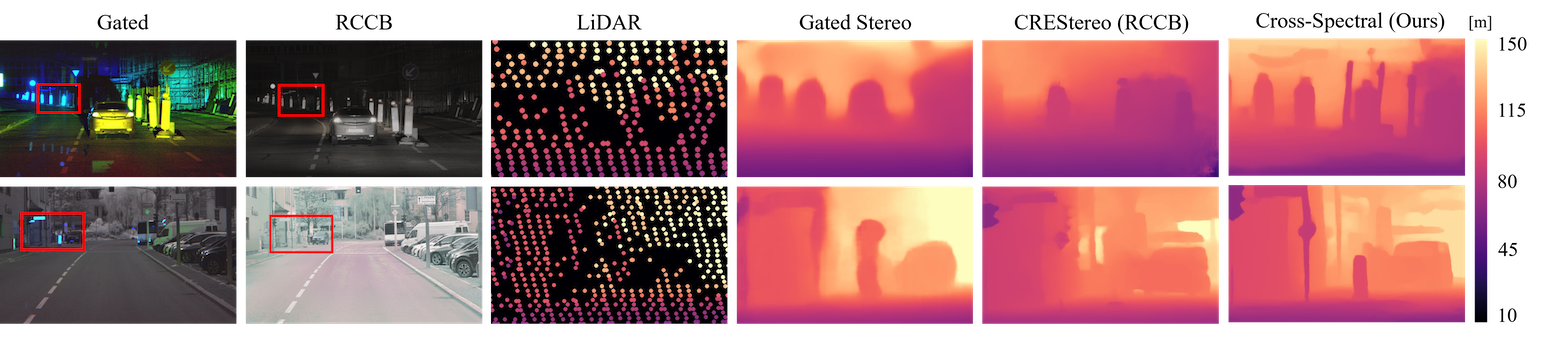}\vspace{-0.5eM}
	\caption{Comparison of our method to LiDAR and the best state-of-the-art methods that rely only on a single modality: Gated Stereo (gated images) \cite{gatedstereo} and CREStereo (RCCB images) \cite{liPracticalStereoMatching2022}. Our method recovers fine details of distant objects irrespective of daylight and nighttime. Limitation of depth range in colorized depth maps for visualization purposes only.}
   \label{fig:zoom_in}
\end{figure*}

\begin{figure*}[!t]
    \vspace*{-3mm}
	\centering
	\includegraphics[width=\textwidth]{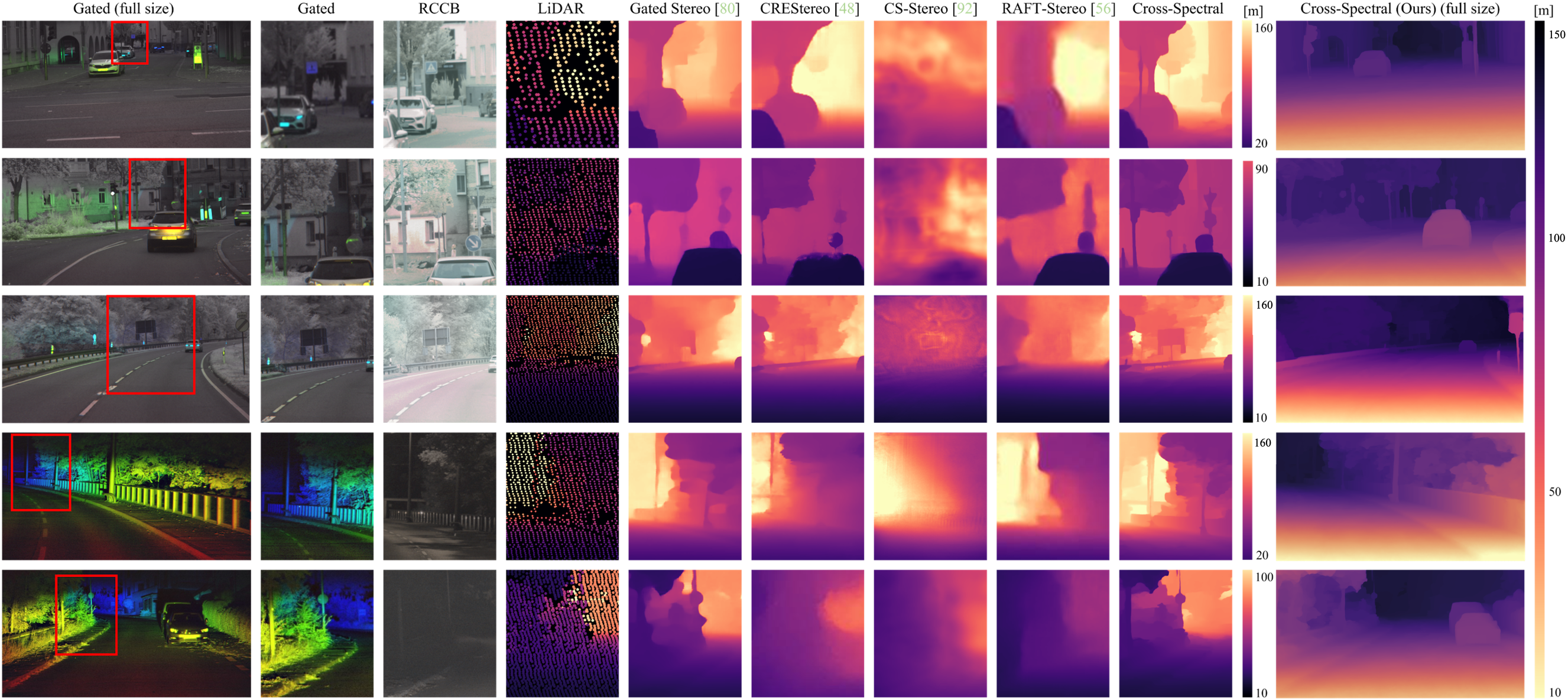}\vspace{-0.9eM}
	\caption{Qualitative comparison of our Gated-RCCB Stereo and existing methods. Our approach is unique in its ability to produce consistently accurate and high-detail depth maps regardless of the ambient illumination condition. In our depth maps, fine structures such as trees or poles are clearly visible, unlike other methods that struggle with consistent depth prediction for these elements. For enhanced visibility of distant objects, the color maps used in zoom-ins are inverted and scaled.}
   \label{fig:comp_ref_methods}
	\vspace*{-0.5cm }
\end{figure*}

\vspace{-2mm}
\PAR{Depth Reconstruction.}
Qualitative results are presented in Figure~\ref{fig:comp_ref_methods} and quantitative results in Table~\ref{tab:results_g2d}. Here, we compare against three recent gated \cite{gated2depth2019,gated2gated, gatedstereo}, six monocular RGB \cite{guizilini20203d,godard2019digging,li2022simipu,bhat2021adabins,ranftl2021vision,li2022depthformer}, six stereo RGB \cite{Chang2018,li2021revisiting,yang2019hsm,xu2022attention,lipson2021raft, liPracticalStereoMatching2022} and two cross-spectral stereo \cite{zhiDeepMaterialAwareCrossSpectral2018, liangUnsupervisedCrossspectralStereo2019} methods. 
Compared to the next best stereo method, Gated Stereo~\cite{gatedstereo}, our method reduces the error by \unit[9.7]{\%} and \unit[0.22]{m} in Mean Absolute Error (MAE) during nighttime conditions and by \unit[13.3]{\%} and \unit[0.3]{m} during day conditions. Additionally, we compare our method to the two next-best stereo methods \cite{gatedstereo, liPracticalStereoMatching2022} on accumulated LiDAR ground-truth maps which allow assessment up to 220 m in Table~\ref{tab:results_g2d_acc}. Our method reduces the error of the next best method averaged over day and night, Gated Stereo~\cite{gatedstereo}, by \unit[16.1]{\%} and \unit[1.4]{m} and CREStereo~\cite{liPracticalStereoMatching2022} by \unit[17.1]{\%} and \unit[1.7]{m}. For distances between 100 and 220 m, our method achieves an improvement of \unit[39.6]{\%} over \cite{gatedstereo} and \unit[39.2]{\%} over \cite{liPracticalStereoMatching2022}, demonstrating a considerable improvement at long distances. Note that \cite{gatedstereo} is designed for distances up to 160 m only. 
Qualitatively, this improvement is visible in sharper edges and rendering of fine details missed by other methods. Compared to the two next-best methods \cite{gatedstereo, liPracticalStereoMatching2022}, the benefits of our cross-spectral depth estimation are highlighted for fine structures at large distances, see Fig.~\ref{fig:zoom_in}. Compared to alternative cross-spectral stereo methods like CS-Stereo \cite{zhiDeepMaterialAwareCrossSpectral2018}, our method is visually and quantitatively superior by a wide margin of \unit[83.0]{\%} as these methods generally don't display details.

\PAR{Qualitative Assessment of Lost Cargo Data}
Our study includes a qualitative comparison of depth estimation methods, focusing on detecting small, potentially hazardous highway objects (as shown in Figure~\ref{fig:lost_cargo}, with more examples in the Supplemental Material). This is critical for autonomous driving, where early detection of such "lost cargo" is necessary for safe maneuvering. Traditional LiDAR often lacks the necessary depth detail for small objects, while passive RCCB cameras are effective in daylight but less so in low light. Gated cameras, although useful, struggle in bright conditions and have lower resolution. Our analysis highlights that high-resolution RCCB data and precise time-of-flight gated data combined with our cross-spectral gated stereo surpass single-modality sensors in detecting small objects at long distances, an essential capability for advanced autonomous driving systems.

\PAR{Ablation Experiments.}
Next, we evaluate the effectiveness of our method by progressively removing components from the full model, see Table~\ref{tab:ablation}.
We start with the full model, achieving the overall best metrics averaged over day and night. First, we downsample the RCCB image to a third of its original height and width, effectively setting the resolution of the depth map to be similar to the resolution of the gated image. This leads to a visible reduction of details in the depth map which cannot be measured quantitatively using sparse LiDAR. To assess the effectiveness of our cross-spectral (CS) training approach, encompassing alternating training, self-supervised losses, and dense LiDAR supervision, we then remove this component, training with sparse LiDAR supervision only, which results in an increase in MAE by 8.4\%. 
Further simplification involves omitting the pose-refinement step within our proposed CSM. This step, too, causes an increase in MAE by 2.1\%, indicating the effectiveness of these components in our method. Subsequent removal of the attention-based feature fusion mechanism and replacing MPViT backbone with the backbone from \cite{liPracticalStereoMatching2022} shows an additional decrease in MAE by 7.3\% and 9.9\%, respectively. Next, we analyze the impact of the dual-camera setup, comprising one RCCB and one gated camera. Discarding this setup leads to more than double the MAE, highlighting the importance of the double stereo camera configuration. 
Finally, we revert to a monocular depth estimation baseline, which records the highest daytime MAE, highlighting the value of stereo cues.

\vspace{-2mm}
\section{Conclusion}
\vspace{-2mm}
In this study, we devise a novel cross-spectral method for stereo depth estimation, combining active gated NIR and high-resolution HDR RCCB cameras. This approach outperforms existing LiDAR sensors in spatial resolution without compromising depth accuracy. Our method is effective in varying lighting conditions, with gated NIR excelling at night and RCCB cameras in daylight. To combine both modalities, we propose a stereo depth estimation method that hinges on a new cross-spectral fusion module trained both supervised and self-supervised losses. Economically viable, our system employs cost-effective CMOS sensors, achieving depth with superior accuracy and quality, surpassing existing methods that rely on single modalities by 39\% in MAE at long ranges. This enables novel applications like long-distance detection of small ground-level objects.

\vspace{-1mm}
\PAR{Acknowledgments} This work was supported by the AI-SEE project with funding from the FFG, BMBF, and NRC-IRA. Felix Heide was supported by an NSF CAREER Award (2047359), a Packard Foundation
Fellowship, a Sloan Research Fellowship, a Sony Young Faculty Award, a Project X Innovation Award, and an Amazon Science Research Award.

{\small
\bibliographystyle{ieee_fullname}
\interlinepenalty=10000
\bibliography{refs}

\begin{thebibliography}{10}\itemsep=-1pt

\bibitem{adam2017bayesian}
Amit Adam, Christoph Dann, Omer Yair, Shai Mazor, and Sebastian Nowozin.
\newblock Bayesian time-of-flight for realtime shape, illumination and albedo.
\newblock {\em IEEE Transactions on Pattern Analysis and Machine Intelligence},
  39(5):851--864, 2017.

\bibitem{aleotti2021neural}
Filippo Aleotti, Fabio Tosi, Pierluigi~Zama Ramirez, Matteo Poggi, Samuele
  Salti, Stefano Mattoccia, and Luigi Di~Stefano.
\newblock Neural disparity refinement for arbitrary resolution stereo.
\newblock In {\em 2021 International Conference on 3D Vision (3DV)}, pages
  207--217. IEEE, 2021.

\bibitem{Andersson2006}
Pierre Andersson.
\newblock Long-range three-dimensional imaging using range-gated laser radar
  images.
\newblock {\em Optical Engineering}, 45(3):034301, 2006.

\bibitem{badki2020Bi3D}
Abhishek Badki, Alejandro Troccoli, Kihwan Kim, Jan Kautz, Pradeep Sen, and
  Orazio Gallo.
\newblock {Bi3D}: {S}tereo depth estimation via binary classifications.
\newblock In {\em arXiv preprint arXiv:2005.07274}, 2020.

\bibitem{bhat2021adabins}
Shariq~Farooq Bhat, Ibraheem Alhashim, and Peter Wonka.
\newblock Adabins: Depth estimation using adaptive bins.
\newblock In {\em Proceedings of the IEEE/CVF Conference on Computer Vision and
  Pattern Recognition}, pages 4009--4018, 2021.

\bibitem{BenchmarkLidar}
Mario Bijelic, Tobias Gruber, and Werner Ritter.
\newblock A benchmark for lidar sensors in fog: Is detection breaking down?
\newblock In {\em 2018 IEEE Intelligent Vehicles Symposium (IV)}, pages
  760--767, 2018.

\bibitem{Bijelic2018}
Mario Bijelic, Tobias Gruber, and Werner Ritter.
\newblock Benchmarking image sensors under adverse weather conditions for
  autonomous driving.
\newblock In {\em IEEE Intelligent Vehicle Symposium}, 2018.

\bibitem{bourlai2012study}
Thirimachos Bourlai, Arun Ross, Cunjian Chen, and Lawrence Hornak.
\newblock A study on using mid-wave infrared images for face recognition.
\newblock In {\em Sensing Technologies for Global Health, Military Medicine,
  Disaster Response, and Environmental Monitoring II; and Biometric Technology
  for Human Identification IX}, volume 8371, pages 239--251. SPIE, 2012.

\bibitem{brown2011multi}
Matthew Brown and Sabine S{\"u}sstrunk.
\newblock {Multi-spectral} sift for scene category recognition.
\newblock In {\em CVPR 2011}, pages 177--184. IEEE, 2011.

\bibitem{Busck2005}
Jens Busck.
\newblock Underwater {3-D} optical imaging with a gated viewing laser radar.
\newblock {\em Optical Engineering}, 2005.

\bibitem{Busck2004}
Jens Busck and Henning Heiselberg.
\newblock Gated viewing and high-accuracy three-dimensional laser radar.
\newblock {\em Applied Optics}, 43(24):4705--10, 2004.

\bibitem{LIBRE}
A. {Carballo}, J. {Lambert}, A. {Monrroy}, D. {Wong}, P. {Narksri}, Y.
  {Kitsukawa}, E. {Takeuchi}, S. {Kato}, and K. {Takeda}.
\newblock Libre: The multiple 3d lidar dataset.
\newblock In {\em IEEE Intelligent Vehicles Symposium (IV)}, 2020.

\bibitem{chang2018pyramid}
Jia-Ren Chang and Yong-Sheng Chen.
\newblock Pyramid stereo matching network.
\newblock In {\em Proceedings of the IEEE Conference on Computer Vision and
  Pattern Recognition}, pages 5410--5418, 2018.

\bibitem{Chang2018}
Jia-Ren Chang and Yong-Sheng Chen.
\newblock Pyramid stereo matching network.
\newblock In {\em Proceedings of the IEEE Conference on Computer Vision and
  Pattern Recognition}, pages 5410--5418, 2018.

\bibitem{ImprovingKinect}
Wei-Chen Chiu, Ulf Blanke, and Mario Fritz.
\newblock Improving the kinect by cross-modal stereo.
\newblock 01 2011.

\bibitem{VolPropagationNetStereoLidar}
Jaesung Choe, Kyungdon Joo, Tooba Imtiaz, and In~So Kweon.
\newblock Volumetric propagation network: Stereo-lidar fusion for long-range
  depth estimation.
\newblock {\em IEEE Robotics and Automation Letters}, 6(3):4672--4679, 2021.

\bibitem{dai2020self}
Qi Dai, Vaishakh Patil, Simon Hecker, Dengxin Dai, Luc Van~Gool, and Konrad
  Schindler.
\newblock Self-supervised object motion and depth estimation from video.
\newblock In {\em Proceedings of the IEEE/CVF Conference on Computer Vision and
  Pattern Recognition Workshops}, pages 1004--1005, 2020.

\bibitem{dai2021attentional}
Yimian Dai, Fabian Gieseke, Stefan Oehmcke, Yiquan Wu, and Kobus Barnard.
\newblock Attentional feature fusion.
\newblock In {\em Proceedings of the IEEE/CVF winter conference on applications
  of computer vision}, pages 3560--3569, 2021.

\bibitem{eigen2014depth}
David Eigen, Christian Puhrsch, and Rob Fergus.
\newblock Depth map prediction from a single image using a multi-scale deep
  network.
\newblock In {\em Advances in Neural Information Processing Systems}, pages
  2366--2374, 2014.

\bibitem{Garg2016}
Ravi Garg, B.G.~Vijay Kumar, Gustavo Carneiro, and Ian Reid.
\newblock Unsupervised {CNN} for single view depth estimation: Geometry to the
  rescue.
\newblock In {\em Proceedings of the IEEE European Conf. on Computer Vision},
  pages 740--756, 2016.

\bibitem{godard2017unsupervised}
Cl{\'e}ment Godard, Oisin Mac~Aodha, and Gabriel~J Brostow.
\newblock Unsupervised monocular depth estimation with left-right consistency.
\newblock In {\em Proceedings of the IEEE conference on computer vision and
  pattern recognition}, pages 270--279, 2017.

\bibitem{godard2019digging}
Cl{\'e}ment Godard, Oisin Mac~Aodha, Michael Firman, and Gabriel~J Brostow.
\newblock Digging into self-supervised monocular depth estimation.
\newblock In {\em Proceedings of the IEEE/CVF International Conference on
  Computer Vision}, pages 3828--3838, 2019.

\bibitem{grauer2014active}
Yoav Grauer.
\newblock Active gated imaging in driver assistance system.
\newblock {\em Advanced Optical Technologies}, 3(2):151--160, 2014.

\bibitem{gated2depth2019}
Tobias Gruber, Frank Julca-Aguilar, Mario Bijelic, and Felix Heide.
\newblock Gated2depth: Real-time dense lidar from gated images.
\newblock In {\em The IEEE International Conference on Computer Vision (ICCV)},
  2019.

\bibitem{gruber2018learning}
Tobias Gruber, Mariia Kokhova, Werner Ritter, Norbert Haala, and Klaus
  Dictmayer.
\newblock Learning super-resolved depth from active gated imaging.
\newblock In {\em 2018 21st International Conference on Intelligent
  Transportation Systems (ITSC)}, pages 3051--3058. IEEE, 2018.

\bibitem{guizilini20203d}
Vitor Guizilini, Rares Ambrus, Sudeep Pillai, Allan Raventos, and Adrien
  Gaidon.
\newblock 3d packing for self-supervised monocular depth estimation.
\newblock In {\em Proceedings of the IEEE/CVF Conference on Computer Vision and
  Pattern Recognition}, pages 2485--2494, 2020.

\bibitem{hansard2012time}
Miles Hansard, Seungkyu Lee, Ouk Choi, and Radu~Patrice Horaud.
\newblock {\em Time-of-flight cameras: principles, methods and applications}.
\newblock Springer Science \& Business Media, 2012.

\bibitem{he2017learning}
Ran He, Xiang Wu, Zhenan Sun, and Tieniu Tan.
\newblock Learning invariant deep representation for nir-vis face recognition.
\newblock In {\em Proceedings of the AAAI Conference on Artificial
  Intelligence}, volume~31, 2017.

\bibitem{heckman1967}
Paul Heckman and Robert~T. Hodgson.
\newblock Underwater optical range gating.
\newblock {\em IEEE Journal of Quantum Electronics}, 3(11):445--448, 1967.

\bibitem{RobustStereo2011}
Yong~Seok Heo, Kyong~Mu Lee, and Sang~Uk Lee.
\newblock Robust stereo matching using adaptive normalized cross-correlation.
\newblock {\em IEEE Transactions on Pattern Analysis and Machine Intelligence},
  33(4):807--822, 2011.

\bibitem{hu2020PENetRGBLidar}
Mu Hu, Shuling Wang, Bin Li, Shiyu Ning, Li Fan, and Xiaojin Gong.
\newblock Towards precise and efficient image guided depth completion.
\newblock 2021.

\bibitem{hu2023planning}
Yihan Hu, Jiazhi Yang, Li Chen, Keyu Li, Chonghao Sima, Xizhou Zhu, Siqi Chai,
  Senyao Du, Tianwei Lin, Wenhai Wang, et~al.
\newblock Planning-oriented autonomous driving.
\newblock In {\em Proceedings of the IEEE/CVF Conference on Computer Vision and
  Pattern Recognition}, pages 17853--17862, 2023.

\bibitem{hwang2015multispectral}
Soonmin Hwang, Jaesik Park, Namil Kim, Yukyung Choi, and In So~Kweon.
\newblock Multispectral pedestrian detection: Benchmark dataset and baseline.
\newblock In {\em Proceedings of the IEEE conference on computer vision and
  pattern recognition}, pages 1037--1045, 2015.

\bibitem{jaritz2018sparse}
Maximilian Jaritz, Raoul De~Charette, Emilie Wirbel, Xavier Perrotton, and
  Fawzi Nashashibi.
\newblock Sparse and dense data with cnns: Depth completion and semantic
  segmentation.
\newblock In {\em International Conference on 3D Vision (3DV)}, pages 52--60,
  2018.

\bibitem{MonochromeRGBStereo2016}
Hae-Gon Jeon, Joon-Young Lee, Sunghoon Im, Hyowon Ha, and In~So Kweon.
\newblock Stereo matching with color and monochrome cameras in low-light
  conditions.
\newblock In {\em 2016 IEEE Conference on Computer Vision and Pattern
  Recognition (CVPR)}, pages 4086--4094, 2016.

\bibitem{Jokela}
Maria Jokela, Matti Kutila, and Pasi Pyykönen.
\newblock Testing and validation of automotive point-cloud sensors in adverse
  weather conditions.
\newblock {\em Applied Sciences}, 9, 2019.

\bibitem{juefei2015nir}
Felix Juefei-Xu, Dipan~K Pal, and Marios Savvides.
\newblock Nir-vis heterogeneous face recognition via cross-spectral joint
  dictionary learning and reconstruction.
\newblock In {\em Proceedings of the IEEE conference on computer vision and
  pattern recognition workshops}, pages 141--150, 2015.

\bibitem{Kendall2017}
Alex Kendall, Hayk Martirosyan, Saumitro Dasgupta, Peter Henry, Ryan Kennedy,
  Abraham Bachrach, and Adam Bry.
\newblock End-to-end learning of geometry and context for deep stereo
  regression.
\newblock In {\em Proceedings of the IEEE International Conference on Computer
  Vision}, 2017.

\bibitem{TPAMODenseCrossModal2021}
Seungryong Kim, Dongbo Min, Stephen Lin, and Kwanghoon Sohn.
\newblock Dense cross-modal correspondence estimation with the deep
  self-correlation descriptor.
\newblock {\em IEEE Transactions on Pattern Analysis and Machine Intelligence},
  43(7):2345--2359, 2021.

\bibitem{kolb2010time}
Andreas Kolb, Erhardt Barth, Reinhard Koch, and Rasmus Larsen.
\newblock Time-of-flight cameras in computer graphics.
\newblock In {\em Computer Graphics Forum}, volume~29, pages 141--159. Wiley
  Online Library, 2010.

\bibitem{lange00tof}
Robert Lange.
\newblock {3D} time-of-flight distance measurement with custom solid-state
  image sensors in {CMOS}/{CCD}-technology.
\newblock 2000.

\bibitem{Laurenzis2009}
Martin Laurenzis, Frank Christnacher, Nicolas Metzger, Emmanuel Bacher, and
  Ingo Zielenski.
\newblock Three-dimensional range-gated imaging at infrared wavelengths with
  super-resolution depth mapping.
\newblock In {\em SPIE Infrared Technology and Applications XXXV}, volume 7298,
  2009.

\bibitem{Laurenzis2007}
Martin Laurenzis, Frank Christnacher, and David Monnin.
\newblock Long-range three-dimensional active imaging with superresolution
  depth mapping.
\newblock {\em Optics letters}, 32(21):3146--8, 2007.

\bibitem{le2020edge20}
Ha Le, Christos Smailis, Lei Shi, and Ioannis Kakadiaris.
\newblock Edge20: A cross spectral evaluation dataset for multiple surveillance
  problems.
\newblock In {\em Proceedings of the IEEE/CVF Winter Conference on Applications
  of Computer Vision}, pages 2685--2694, 2020.

\bibitem{lee2022mpvit}
Youngwan Lee, Jonghee Kim, Jeffrey Willette, and Sung~Ju Hwang.
\newblock Mpvit: Multi-path vision transformer for dense prediction.
\newblock In {\em Proceedings of the IEEE/CVF Conference on Computer Vision and
  Pattern Recognition}, pages 7287--7296, 2022.

\bibitem{Lezama_2017_CVPR}
Jose Lezama, Qiang Qiu, and Guillermo Sapiro.
\newblock Not afraid of the dark: Nir-vis face recognition via cross-spectral
  hallucination and low-rank embedding.
\newblock In {\em Proceedings of the IEEE Conference on Computer Vision and
  Pattern Recognition (CVPR)}, July 2017.

\bibitem{li2023mseg3d}
Jiale Li, Hang Dai, Hao Han, and Yong Ding.
\newblock Mseg3d: Multi-modal 3d semantic segmentation for autonomous driving.
\newblock In {\em Proceedings of the IEEE/CVF Conference on Computer Vision and
  Pattern Recognition}, pages 21694--21704, 2023.

\bibitem{liPracticalStereoMatching2022}
Jiankun Li, Peisen Wang, Pengfei Xiong, Tao Cai, Ziwei Yan, Lei Yang, Jiangyu
  Liu, Haoqiang Fan, and Shuaicheng Liu.
\newblock Practical {{Stereo Matching}} via {{Cascaded Recurrent Network}} with
  {{Adaptive Correlation}}.
\newblock In {\em 2022 {{IEEE}}/{{CVF Conference}} on {{Computer Vision}} and
  {{Pattern Recognition}} ({{CVPR}})}, pages 16242--16251, {New Orleans, LA,
  USA}, Jun 2022. {IEEE}.

\bibitem{Li_2013_CVPR_Workshops}
Stan~Z. Li, Dong Yi, Zhen Lei, and Shengcai Liao.
\newblock The casia nir-vis 2.0 face database.
\newblock In {\em Proceedings of the IEEE Conference on Computer Vision and
  Pattern Recognition (CVPR) Workshops}, June 2013.

\bibitem{li2022simipu}
Zhenyu Li, Zehui Chen, Ang Li, Liangji Fang, Qinhong Jiang, Xianming Liu,
  Junjun Jiang, Bolei Zhou, and Hang Zhao.
\newblock Simipu: Simple 2d image and 3d point cloud unsupervised pre-training
  for spatial-aware visual representations.
\newblock In {\em Proceedings of the AAAI Conference on Artificial
  Intelligence}, volume~36, pages 1500--1508, 2022.

\bibitem{li2022depthformer}
Zhenyu Li, Zehui Chen, Xianming Liu, and Junjun Jiang.
\newblock Depthformer: Depthformer: Exploiting long-range correlation and local
  information for accurate monocular depth estimation.
\newblock {\em arXiv preprint arXiv:2203.14211}, 2022.

\bibitem{MovingPeopleMovingCameras}
Zhengqi Li, Tali Dekel, Forrester Cole, Richard Tucker, Noah Snavely, Ce Liu,
  and William~T. Freeman.
\newblock Mannequinchallenge: Learning the depths of moving people by watching
  frozen people.
\newblock {\em IEEE Transactions on Pattern Analysis and Machine Intelligence},
  43(12):4229--4241, 2021.

\bibitem{TransformerStereo}
Zhaoshuo Li, Xingtong Liu, Nathan Drenkow, Andy Ding, Francis~X. Creighton,
  Russell~H. Taylor, and Mathias Unberath.
\newblock Revisiting stereo depth estimation from a sequence-to-sequence
  perspective with transformers.
\newblock In {\em 2021 IEEE/CVF International Conference on Computer Vision
  (ICCV)}, pages 6177--6186, 2021.

\bibitem{li2021revisiting}
Zhaoshuo Li, Xingtong Liu, Nathan Drenkow, Andy Ding, Francis~X Creighton,
  Russell~H Taylor, and Mathias Unberath.
\newblock Revisiting stereo depth estimation from a sequence-to-sequence
  perspective with transformers.
\newblock In {\em Proceedings of the IEEE/CVF International Conference on
  Computer Vision}, pages 6197--6206, 2021.

\bibitem{liangUnsupervisedCrossspectralStereo2019}
Mingyang Liang, Xiaoyang Guo, Hongsheng Li, Xiaogang Wang, and You Song.
\newblock Unsupervised {{Cross-spectral Stereo Matching}} by {{Learning}} to
  {{Synthesize}}, Mar 2019.

\bibitem{lipson2021raft}
Lahav Lipson, Zachary Teed, and Jia Deng.
\newblock Raft-stereo: Multilevel recurrent field transforms for stereo
  matching.
\newblock In {\em 2021 International Conference on 3D Vision (3DV)}, pages
  218--227. IEEE, 2021.

\bibitem{luo2019every}
Chenxu Luo, Zhenheng Yang, Peng Wang, Yang Wang, Wei Xu, Ram Nevatia, and Alan
  Yuille.
\newblock Every pixel counts++: Joint learning of geometry and motion with 3d
  holistic understanding.
\newblock {\em IEEE transactions on pattern analysis and machine intelligence},
  42(10):2624--2641, 2019.

\bibitem{ma2018sparse}
Fangchang Ma and Sertac Karaman.
\newblock Sparse-to-dense: Depth prediction from sparse depth samples and a
  single image.
\newblock In {\em IEEE International Conference on Robotics and Automation},
  pages 1--8, 2018.

\bibitem{Mayer2016}
N. Mayer, E. Ilg, P. H{\"a}usser, P. Fischer, D. Cremers, A. Dosovitskiy, and
  T. Brox.
\newblock A large dataset to train convolutional networks for disparity,
  optical flow, and scene flow estimation.
\newblock In {\em Proceedings of the IEEE Conference on Computer Vision and
  Pattern Recognition}, 2016.

\bibitem{miangoleh2021boosting}
S~Mahdi~H Miangoleh, Sebastian Dille, Long Mai, Sylvain Paris, and Yagiz Aksoy.
\newblock Boosting monocular depth estimation models to high-resolution via
  content-adaptive multi-resolution merging.
\newblock In {\em Proceedings of the IEEE/CVF Conference on Computer Vision and
  Pattern Recognition}, pages 9685--9694, 2021.

\bibitem{park2020nonRGBLidar}
Jinsun Park, Kyungdon Joo, Zhe Hu, Chi-Kuei Liu, and In~So Kweon.
\newblock Non-local spatial propagation network for depth completion.
\newblock In {\em Proc. of European Conference on Computer Vision (ECCV)},
  2020.

\bibitem{Pinggera2012OnCS}
Peter Pinggera, T. Breckon, and Horst Bischof.
\newblock On cross-spectral stereo matching using dense gradient features.
\newblock In {\em British Machine Vision Conference}, 2012.

\bibitem{qiaoDepthSuperResolutionExplicit2023}
Xin Qiao, Chenyang Ge, Youmin Zhang, Yanhui Zhou, Fabio Tosi, Matteo Poggi, and
  Stefano Mattoccia.
\newblock Depth {{Super-Resolution}} from {{Explicit}} and {{Implicit
  High-Frequency Features}}, May 2023.

\bibitem{ranftl2021vision}
Ren{\'e} Ranftl, Alexey Bochkovskiy, and Vladlen Koltun.
\newblock Vision transformers for dense prediction.
\newblock In {\em Proceedings of the IEEE/CVF International Conference on
  Computer Vision}, pages 12179--12188, 2021.

\bibitem{ranjan2019competitive}
Anurag Ranjan, Varun Jampani, Lukas Balles, Kihwan Kim, Deqing Sun, Jonas
  Wulff, and Michael~J Black.
\newblock Competitive collaboration: Joint unsupervised learning of depth,
  camera motion, optical flow and motion segmentation.
\newblock In {\em Proceedings of the IEEE/CVF Conference on Computer Vision and
  Pattern Recognition}, pages 12240--12249, 2019.

\bibitem{rufenacht2013automatic}
Dominic R{\"u}fenacht, Cl{\'e}ment Fredembach, and Sabine S{\"u}sstrunk.
\newblock Automatic and accurate shadow detection using near-infrared
  information.
\newblock {\em IEEE transactions on pattern analysis and machine intelligence},
  36(8):1672--1678, 2013.

\bibitem{schober2017dynamic}
Michael Schober, Amit Adam, Omer Yair, Shai Mazor, and Sebastian Nowozin.
\newblock Dynamic time-of-flight.
\newblock In {\em Proceedings of the IEEE Conference on Computer Vision and
  Pattern Recognition}, pages 6109--6118, 2017.

\bibitem{schwarz2010lidar}
Brent Schwarz.
\newblock Lidar: Mapping the world in {3D}.
\newblock {\em Nature Photonics}, 4(7):429, 2010.

\bibitem{shan2020lio}
Tixiao Shan, Brendan Englot, Drew Meyers, Wei Wang, Carlo Ratti, and Daniela
  Rus.
\newblock Lio-sam: Tightly-coupled lidar inertial odometry via smoothing and
  mapping.
\newblock In {\em 2020 IEEE/RSJ international conference on intelligent robots
  and systems (IROS)}, pages 5135--5142. IEEE, 2020.

\bibitem{MultiSpectral2014}
Xiaoyong Shen, Li Xu, Qi Zhang, and Jiaya Jia.
\newblock Multi-modal and multi-spectral registration for natural images.
\newblock In David Fleet, Tomas Pajdla, Bernt Schiele, and Tinne Tuytelaars,
  editors, {\em Computer Vision -- ECCV 2014}, pages 309--324, Cham, 2014.
  Springer International Publishing.

\bibitem{WaymoDataset}
Pei Sun, Henrik Kretzschmar, Xerxes Dotiwalla, Aurelien Chouard, Vijaysai
  Patnaik, Paul Tsui, James Guo, Yin Zhou, Yuning Chai, Benjamin Caine, Vijay
  Vasudevan, Wei Han, Jiquan Ngiam, Hang Zhao, Aleksei Timofeev, Scott
  Ettinger, Maxim Krivokon, Amy Gao, Aditya Joshi, Yu Zhang, Jonathon Shlens,
  Zhifeng Chen, and Dragomir Anguelov.
\newblock Scalability in perception for autonomous driving: Waymo open dataset.
\newblock In {\em Proceedings of the IEEE/CVF Conference on Computer Vision and
  Pattern Recognition (CVPR)}, June 2020.

\bibitem{tang2019sparse2dense}
Jiexiong Tang, John Folkesson, and Patric Jensfelt.
\newblock Sparse2dense: From direct sparse odometry to dense 3-d
  reconstruction.
\newblock {\em IEEE Robotics and Automation Letters}, 4(2):530--537, 2019.

\bibitem{guidenetRGBLidar}
Jie Tang, Fei-Peng Tian, Wei Feng, Jian Li, and Ping Tan.
\newblock Learning guided convolutional network for depth completion.
\newblock {\em IEEE Transactions on Image Processing}, 30:1116--1129, 2020.

\bibitem{Thavalengal_2015_CVPR_Workshops}
Shejin Thavalengal, Petronel Bigioi, and Peter Corcoran.
\newblock Evaluation of combined visible/nir camera for iris authentication on
  smartphones.
\newblock In {\em Proceedings of the IEEE Conference on Computer Vision and
  Pattern Recognition (CVPR) Workshops}, June 2015.

\bibitem{tosiRGBMultispectralMatchingDataset2022a}
Fabio Tosi, Pierluigi~Zama Ramirez, Matteo Poggi, Samuele Salti, Stefano
  Mattoccia, and Luigi Di~Stefano.
\newblock {{RGB-Multispectral Matching}}: {{Dataset}}, {{Learning
  Methodology}}, {{Evaluation}}.
\newblock In {\em 2022 {{IEEE}}/{{CVF Conference}} on {{Computer Vision}} and
  {{Pattern Recognition}} ({{CVPR}})}, pages 15937--15947, {New Orleans, LA,
  USA}, Jun 2022. {IEEE}.

\bibitem{vijayanarasimhan2017sfm}
Sudheendra Vijayanarasimhan, Susanna Ricco, Cordelia Schmid, Rahul Sukthankar,
  and Katerina Fragkiadaki.
\newblock Sfm-net: Learning of structure and motion from video.
\newblock {\em arXiv preprint arXiv:1704.07804}, 2017.

\bibitem{villa2012spad}
F Villa, B Markovic, S Bellisai, D Bronzi, A Tosi, F Zappa, S Tisa, D Durini, S
  Weyers, U Paschen, et~al.
\newblock {SPAD} smart pixel for time-of-flight and time-correlated
  single-photon counting measurements.
\newblock {\em IEEE Photonics Journal}, 4(3):795--804, 2012.

\bibitem{gated2gated}
Amanpreet Walia, Stefanie Walz, Mario Bijelic, Fahim Mannan, Frank
  Julca-Aguilar, Michael Langer, Werner Ritter, and Felix Heide.
\newblock Gated2gated: Self-supervised depth estimation from gated images.
\newblock 2022.

\bibitem{waltersThereBackAgain2021}
Celyn Walters, Oscar Mendez, Mark Johnson, and Richard Bowden.
\newblock There and {{Back Again}}: {{Self-supervised Multispectral
  Correspondence Estimation}}.
\newblock In {\em 2021 {{IEEE International Conference}} on {{Robotics}} and
  {{Automation}} ({{ICRA}})}, pages 5147--5154, May 2021.

\bibitem{gatedstereo}
Stefanie Walz, Mario Bijelic, Andrea Ramazzina, Amanpreet Walia, Fahim Mannan,
  and Felix Heide.
\newblock Gated stereo: Joint depth estimation from gated and wide-baseline
  active stereo cues.
\newblock In {\em Proceedings of the IEEE/CVF Conference on Computer Vision and
  Pattern Recognition}, pages 13252--13262, 2023.

\bibitem{mems:mirrors:lidar:review}
Dingkang Wang, Connor Watkins, and Huikai Xie.
\newblock {MEMS} mirrors for {LiDAR}: A review.
\newblock {\em Micromachines}, 11(5), 2020.

\bibitem{PseudoLidar}
Yan Wang, Wei-Lun Chao, Divyansh Garg, Bharath Hariharan, Mark Campbell, and
  Kilian~Q. Weinberger.
\newblock Pseudo-lidar from visual depth estimation: Bridging the gap in 3d
  object detection for autonomous driving.
\newblock In {\em 2019 IEEE/CVF Conference on Computer Vision and Pattern
  Recognition (CVPR)}, pages 8437--8445, 2019.

\bibitem{ssim}
Z. Wang, C Bovik, H.~R. Sheikh, and E.P. Simoncelli.
\newblock Image quality assessment: from error visibility to structural
  similarity.
\newblock In {\em IEEE Transactions on Image Processing}, 2004.

\bibitem{wong2021unsupervised}
Alex Wong and Stefano Soatto.
\newblock Unsupervised depth completion with calibrated backprojection layers.
\newblock In {\em Proceedings of the IEEE/CVF International Conference on
  Computer Vision}, pages 12747--12756, 2021.

\bibitem{Xinwei2013}
Wang Xinwei, Li Youfu, and Zhou Yan.
\newblock Triangular-range-intensity profile spatial-correlation method for
  {3D} super-resolution range-gated imaging.
\newblock {\em Applied Optics}, 52(30):7399--406, 2013.

\bibitem{xu2017learning}
Dan Xu, Wanli Ouyang, Elisa Ricci, Xiaogang Wang, and Nicu Sebe.
\newblock Learning cross-modal deep representations for robust pedestrian
  detection.
\newblock In {\em Proceedings of the IEEE conference on computer vision and
  pattern recognition}, pages 5363--5371, 2017.

\bibitem{xu2022attention}
Gangwei Xu, Junda Cheng, Peng Guo, and Xin Yang.
\newblock Attention concatenation volume for accurate and efficient stereo
  matching.
\newblock In {\em Proceedings of the IEEE/CVF Conference on Computer Vision and
  Pattern Recognition}, pages 12981--12990, 2022.

\bibitem{yang2019hsm}
Gengshan Yang, Joshua Manela, Michael Happold, and Deva Ramanan.
\newblock Hierarchical deep stereo matching on high-resolution images.
\newblock In {\em The IEEE Conference on Computer Vision and Pattern
  Recognition (CVPR)}, June 2019.

\bibitem{yin2018geonet}
Zhichao Yin and Jianping Shi.
\newblock Geonet: Unsupervised learning of dense depth, optical flow and camera
  pose.
\newblock In {\em Proceedings of the IEEE conference on computer vision and
  pattern recognition}, pages 1983--1992, 2018.

\bibitem{SLFNetStereoLidar}
Yongjian Zhang, Longguang Wang, Kunhong Li, Zhiheng Fu, and Yulan Guo.
\newblock Slfnet: A stereo and lidar fusion network for depth completion.
\newblock {\em IEEE Robotics and Automation Letters}, 7(4):10605--10612, 2022.

\bibitem{zhaoSphericalSpaceFeature2023}
Zixiang Zhao, Jiangshe Zhang, Xiang Gu, Chengli Tan, Shuang Xu, Yulun Zhang,
  Radu Timofte, and Luc Van~Gool.
\newblock Spherical {{Space Feature Decomposition}} for {{Guided Depth Map
  Super-Resolution}}, Aug 2023.

\bibitem{zhiDeepMaterialAwareCrossSpectral2018}
Tiancheng Zhi, Bernardo~R Pires, Martial Hebert, and Srinivasa~G Narasimhan.
\newblock Deep material-aware cross-spectral stereo matching.
\newblock In {\em Proceedings of the IEEE conference on computer vision and
  pattern recognition}, pages 1916--1925, 2018.

\bibitem{Zhou2017}
Tinghui Zhou, Matthew Brown, Noah Snavely, and David~G. Lowe.
\newblock Unsupervised learning of depth and ego-motion from video.
\newblock In {\em Proceedings of the IEEE Conference on Computer Vision and
  Pattern Recognition}, 2017.

\end{thebibliography}
}

\end{document}